\documentclass[final]{cvpr}

\usepackage{times}
\usepackage{epsfig}
\usepackage{graphicx}
\usepackage{amsmath}
\usepackage{amssymb}

\usepackage{color}
\usepackage{booktabs} 
\usepackage{bbm}
\usepackage{siunitx}

\usepackage{adjustbox}
\usepackage{chngpage}
\usepackage{multirow}
\usepackage{amssymb}
\usepackage{graphicx}
\usepackage[shortlabels]{enumitem}
\usepackage{array}
\usepackage{tabularx,booktabs}

\usepackage[pagebackref=true,breaklinks=true,colorlinks,bookmarks=false]{hyperref}

\DeclareMathOperator*{\argmin}{arg\,min}

\definecolor{ben}{rgb}{0.9,0.,0.5}

\definecolor{hyun}{rgb}{0.,0.5,0.9}

\definecolor{todo}{rgb}{1.0, 0., 0.}

\def\cvprPaperID{8406} 
\def\confYear{CVPR 2021}
\begin{document}

\title{I Like to Move It: 6D Pose Estimation as an Action Decision Process}

\author{Benjamin Busam
\quad Hyun Jun Jung
\quad Nassir Navab\\
Technical University of Munich\\
{{\tt\small \{b.busam,hyunjun.jung,nassir.navab\}@tum.de}
}}


\maketitle

\begin{abstract}\noindent
Object pose estimation is an integral part of robot vision and AR.
Previous 6D pose retrieval pipelines treat the problem either as a regression task or discretize the pose space to classify.
We change this paradigm and reformulate the problem as an action decision process where an initial pose is updated in incremental discrete steps that sequentially move a virtual 3D rendering towards the correct solution.
A neural network estimates likely moves from a single RGB image iteratively and determines so an acceptable final pose.
In comparison to other approaches that train object-specific pose models, we learn a decision process.
This allows for a lightweight architecture while it naturally generalizes to unseen objects.
A coherent stop action for process termination enables dynamic reduction of the computation cost if there are insignificant changes in a video sequence.
Instead of a static inference time, we thereby automatically increase the runtime depending on the object motion.
Robustness and accuracy of our action decision network are evaluated on Laval~\cite{garon2018framework} and YCB~\cite{xiang2018posecnn} video scenes where we significantly improve the state-of-the-art.
\end{abstract}

\begin{figure}[t]
\centering
\includegraphics[width=\linewidth]{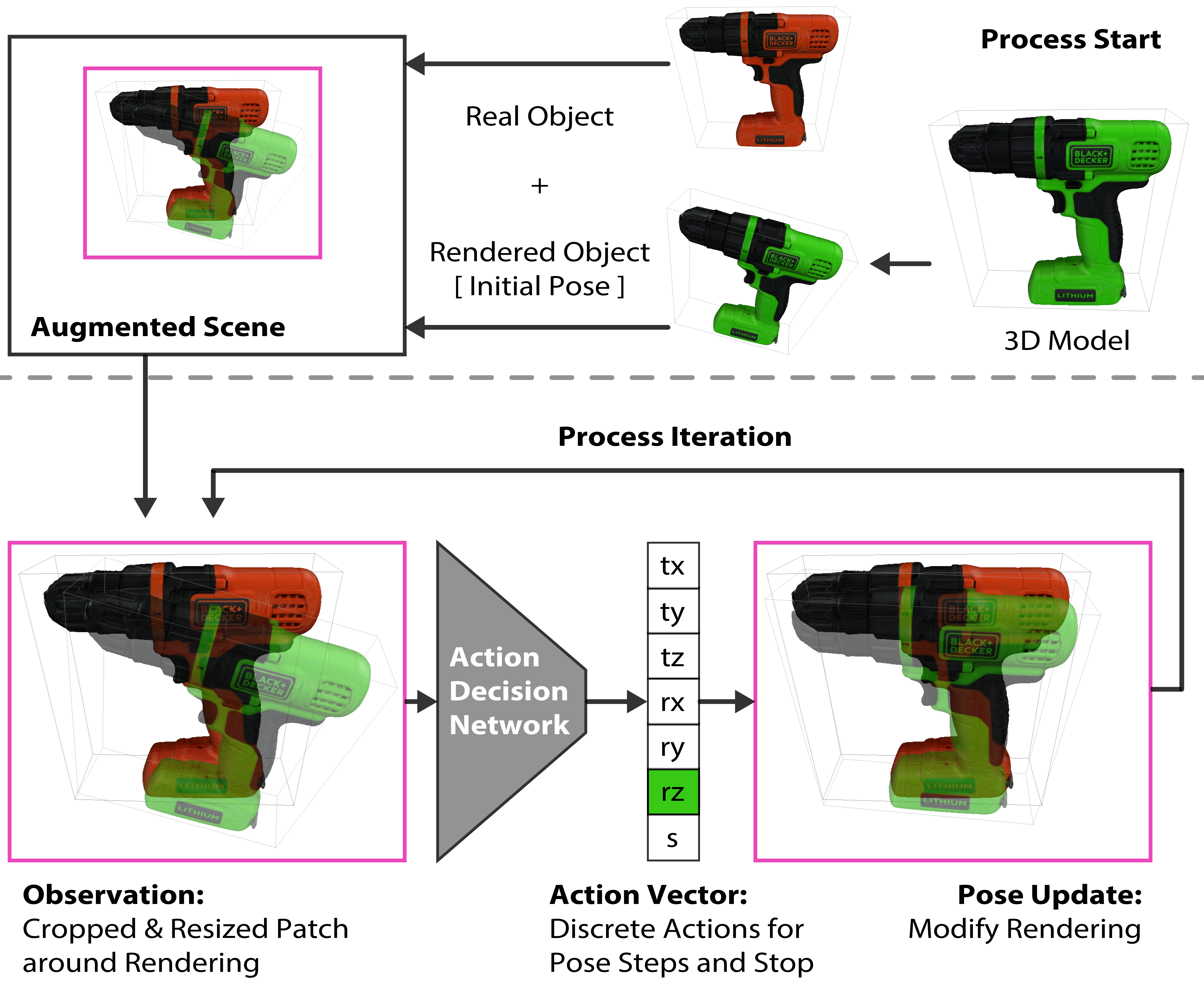}
\caption{\textbf{Method Overview.} A virtual object is rendered with an initial pose on top of a real object (top). Both image and rendering are cropped. A lightweight action decision network determines an incremental move to bring the rendering closer to the real observation. The updated pose is used to iteratively modify the rendering.}
\label{fig:teaser}
\end{figure}

\section{Introduction}\noindent
We live in a 3D world.
Every object with which we interact has six degrees of freedom to move freely in space, three for its orientation and three for its translation.
Thus, the question to determine these parameters naturally arises whenever we include a vision system observing the scene.
A single camera will only capture a projection of this world.
Thus, recovering such information constitutes an inherently ill-posed problem which has drawn attention of many vision experts in the past~\cite{kato1999marker,epnp_2009,hinterstoisser2012,mur2015orb,xiang2018posecnn}.
The motives for this can be different: One may want to extract scene content for accurate measurements~\cite{birdal2016x}, camera localization~\cite{mur2017orb} or 3D reconstruction~\cite{Knapitsch2017}.
Another driver can be geometric image manipulation~\cite{Occlusion2018,busam2019sterefo} or sensor fusion~\cite{esposito2015cooperative}.
Also human-robot interaction~\cite{busam2015stereo} and robot grasping~\cite{drost2017introducing} require estimation of 6D poses.\\
The rise of low-cost RGBD sensors helped development of 6D pose detectors~\cite{brachmann2014,kehl2016deep,wang2019densefusion} and trackers~\cite{Tan_2014_CVPR,garon2018framework}.
More recently, the field also considered methods with single RGB image input.
The best performing methods for this task are all data-driven and thus require a certain amount of training images.
Annotating a large body of data for this kind of task is cumbersome and time-intensive which yields to either complex acquisition setups~\cite{garon2018framework} or diverse annotation quality.
The majority of pose estimation pipelines train on these annotations, usually one network per object.
Besides difficult data annotation and the requirement for annotations for each new object, this brings two further drawbacks.
On one hand, the networks adjust to the individual sensor noise of the acquisition hardware which drastically hampers the generalization capabilities~\cite{Kaskman_2019_ICCV_Workshops}.
On the other hand, every real annotation has its own errors introduced either by the used ground truth sensor system or by the human annotator, which propagate to models trained on it.
Modern 3D renderers, however, can produce photorealistic images in high quantity with pixel-perfect ground truth.
Some recent scholars therefore propose to leverage such data~\cite{kehl2017ssd,sundermeyer2018implicit,Zakharov_2019_ICCV} and fully train on synthetic images.
Most widely used evaluation datasets~\cite{hinterstoisser2012,brachmann2014} provide single image acquisitions and only recently video sequences~\cite{xiang2018posecnn,garon2018framework} with pose annotations are available even though video data is the natural data source in applications.\\

\noindent\textbf{Contributions and Outline.}
We leverage the temporal component in video data to accelerate our pose estimation performance and propose an RGB pose estimation pipeline by taking inspiration from the reinforcement learning approach proposed for 2D bounding box tracking~\cite{Yun_2017_CVPR} where the authors frame the problem with consecutive discrete actions for an agent.
We frame 6D pose estimation as an action decision process realized by applying a network that determines a sequence of likely object moves as shown in Fig.~\ref{fig:teaser}.
At first, an initial pose is used to render the 3D model.
Both the rendering and the current image are cropped around the virtual pose and fed to a lightweight CNN.
The network predicts a pose action to move the rendering closer to the real object.
The pose is then modified according to the action and a new rendering is fed back into the pipeline with a new crop to move the estimation incrementally closer to the observation.
This goes on until either a stop criterion fires or the maximum number of iterations is reached.
If our input is a video stream, we can use the pose retrieved at frame $t$ as an initial pose for frame $t+1$ which can greatly reduce the computation time as the amount of iterations is determined by the pose actions needed between the initial pose and the result.
Improving pose estimation with iterative inference has previously been explored by~\cite{li2018deepim} where a refinement network is iteratively applied to refine a pose predicted by an estimator such as PoseCNN~\cite{xiang2018posecnn}.
However, the performance of their method actually decreases if more than two iterations are used.\\
In summary, our contributions in this work are fourfold:
\begin{enumerate}  
\item We reformulate 6D pose estimation as an \textbf{action decision process} and design a lightweight CNN architecture for this task that \textbf{generalizes to unseen objects}.
\item We \textbf{iteratively} apply a shallow network to optimize the pose and deploy a change-aware \textbf{dynamic complexity reduction} scheme to improve inference cost.
\item We propose an RGB-only method able to \textbf{improve} \textbf{video pose estimation} while being able to track objects in presence of noise and clutter.
\item We provide a \textbf{data augmentation scheme} to render high-quality images of 3D models on real backgrounds under varying clutter and occlusion.
\end{enumerate}
 In the remainder of the paper, we first review the related literature in Section~\ref{sec:related} before we present our method and network architecture(Sec.~\ref{sec:method}).
 An extensive analysis and evaluation of our method is reported in Sec.~\ref{sec:experiments}.
\section{Related Work}
\label{sec:related}\noindent
Vision system acquire imagery of our 3D world.
In order to interact with objects in this world it is crucial to understand relative position and orientation which has been addressed in different ways in the literature.\\
\noindent\textbf{From Markers to Features.}
Early works apply marker based systems to track objects.
Typical augmented reality applications are driven by markers such as AR-Tag~\cite{fiala2005artag}, ArUcO~\cite{garrido2014automatic}, ARToolkit~\cite{kato1999marker} or AprilTag~\cite{olson2011apriltag}.
These are also used for sensor fusion~\cite{esposito2015cooperative} and extended to high accuracy systems~\cite{birdal2016x}.
Reliable and robust detection is of particular interest in the medical domain~\cite{esposito2016multimodal}, where self-adhesive markers allow flexible usage~\cite{busam2015stereo}.\\
Object-marker calibration can be intricate and time-consuming in practice and feature extractors are a practicable alternative.
Methods such as SIFT~\cite{lowe2004distinctive}, SURF~\cite{bay2006surf}, BRISK~\cite{leutenegger2011brisk}, ORB~\cite{orb_2011} etc. are utilized for camera~\cite{mur2015orb,mur2017orb} and object~\cite{wu20083d,li2010location} pose estimation.
Tracking applications benefit from the rotation accuracy of such systems in inside-out camera setups~\cite{busam2018markerless}.
The Perspective-$n$-Point (P$n$P) algorithm and its successor EP$n$P~\cite{epnp_2009} are still utilized to recover 6D poses from 2D-3D correspondences.\\
\noindent\textbf{Pose Regression and Classification.}
Rotations densely populate a non-Euclidean space and there are multiple parametrization for the Riemannian manifold described by them~\cite{busam2016}.
The geodesic distance on the unit quaternions hypersphere is not compliant with the Euclidean L-p norm in its 4D-embedding impeding 6D pose regression networks~\cite{zhou2019continuity}.
Some works therefore discretize the problem and classify~\cite{kehl2017ssd}.
Hinterstoisser~\cite{hinterstoisser2012} uses a template matching strategy for viewpoint estimation and~\cite{cai2013fast,kehl2015hashmod,hodavn2015detection} achieve a sub-linear matching complexity in the number of objects by hashing.\\
Others train a regressor for RGBD~\cite{brachmann2014,tejani2014latent,wohlhart2015learning,kehl2016deep,wang2019densefusion} pose estimation.
Some scholars recently also report methods that solely rely on RGB~\cite{crivellaro2015novel,kehl2017ssd,rad2017bb8,do2018deep,xiang2018posecnn,sundermeyer2018implicit} input without the need of additional depth.
To realize a 6D pose estimation pipeline, these methods are usually separated into three stages~\cite{kehl2017ssd,sundermeyer2018implicit,rad2017bb8}: 2D detection, 2D keypoint extraction, 6D pose estimation.
Tekin~\cite{tekin2018real} is based on YOLO~\cite{redmon2016you} and thus provides a single shot method.
After bounding box corner or keypoint detection, the 6D pose is estimated with P$n$P.
Other approaches~\cite{hu2019segmentation,xiang2018posecnn,do2018deep,pavlakos20176} utilize multi-modalities or multi-task training.
More recently, pixel-wise object correspondences~\cite{Zakharov_2019_ICCV,peng2019pvnet,Li_2019_ICCV} use robust P$n$P within a RANSAC loop to improve the results.
The model performance is mostly hampered by the domain gap created through synthetic-only data training which is addressed for depth renderings by~\cite{rad2018domain}.
Further works address occlusion~\cite{oberweger2018making,fu2019deephmap} and ambiguous~\cite{manhardt2019explaining} cases.
To improve upon the estimated pose, Li et al.~\cite{li2018deepim} propose an RGB-based refinement strategy.
Many methods, however, refine their RGB results with additional depth information using ICP~\cite{zhang1994iterative}.
All the core networks usually require to train one network per object.
If training is done for multiple objects, the resulting predictions become unreliable~\cite{Kaskman_2019_ICCV_Workshops}.
The recent CorNet~\cite{pitteri2019cornet} focuses on objects geometry instead and detects object-agnostic corners.
While this is more robust, it is in spirit similar to early pose estimation approaches that detect significant points.
Our model is different as we learn a discrete set of decisions that gradually lead to the correct pose.
Iterative processes are also used in a reinforcement learning (RL) context for pose estimation with weak supervision by~\cite{shao2020pfrl} while~\cite{krull2017poseagent} uses RL for efficient computation.
Both methods train object-specific networks on real data while we leverage synthetic training.
Some recent works also address joint pose estimation and shape retrieval~\cite{wang2019normalized,manhardt2020cps}. While being instance-agnostic, they are however bound to objects from the same class.

\noindent\textbf{Temporal Tracking.}
Tracking of 3D objects using temporal information has been presented with the help of depth maps and point clouds.
It can be realized with ICP~\cite{besl1992method} and its variants~\cite{rusinkiewicz2001efficient,segal2009generalized}.
These methods highly rely on an initial pose close to the correct prediction and fail in the presence of heavy noise and clutter~\cite{garon2018framework}.
To stabilise tracking with depth maps, additional intensity information~\cite{held20123d,yuheng2013star3d,joseph2015versatile,kehl2017real} or a robust learning procedure~\cite{Tan_2014_CVPR} helps.
The current methods need one CNN trained per objects~\cite{garon2017deep} or are bound to specific geometrical constraints such as planar objects~\cite{wang2017gracker}.
The recent PoseRBPF~\cite{deng2019poserbpf} presents as an efficient RGB-only tracker using a particle filter setting state-of-the-art results on the YCB dataset~\cite{xiang2018posecnn}.
Although our approach may appear similar to a classical temporal tracker whose optimization procedure usually includes incremental pose updates and requires initialization close to the correct pose in order not to fail or drift~\cite{zhang1994iterative}, the convergence basin of our method is much wider (see Sec.~\ref{subsec:robustness}).
While we largely benefit from temporal information in terms of computation time, our method can also be used to detect the pose with multiple seeds intuitively.\\
\noindent\textbf{Pose Datasets.}
To compare different tracking and detection method for 6D pose estimation, different datasets exist.
LineMOD~\cite{hinterstoisser2012} and its occlusion extension~\cite{brachmann2014} are arguably the most widely used ones for detection.
More recently HomebrewedDB~\cite{Kaskman_2019_ICCV_Workshops} uses three of the LineMOD objects and adds 30 higher quality 3D models.
The scenes are more cluttered and acquired under different illumination conditions.
Other datasets focus on textureless industrial~\cite{hodan2017t,drost2017introducing} and household~\cite{rennie2016dataset,doumanoglou2016recovering,tejani2014latent} objects.
A dataset summary is given in the BOP 6D Pose Benchmark~\cite{hodan2018bop}.
While the different setups are diverse and the ground truth labels often of very high quality, objects are usually acquired from individual acquisitions that are not temporally connected making tracking evaluation difficult.
The more recent YCB-Video dataset~\cite{xiang2018posecnn}, however, includes 92 video sequences of 21 household objects and 12 test videos with annotated ground truth poses and detailed 3D models.
Several RGBD trackers also evaluate on the dataset of Garon et al.~\cite{garon2018framework} that includes severe occlusion and clutter.

\begin{figure}[t]
    \centering
    \includegraphics[width=\linewidth]{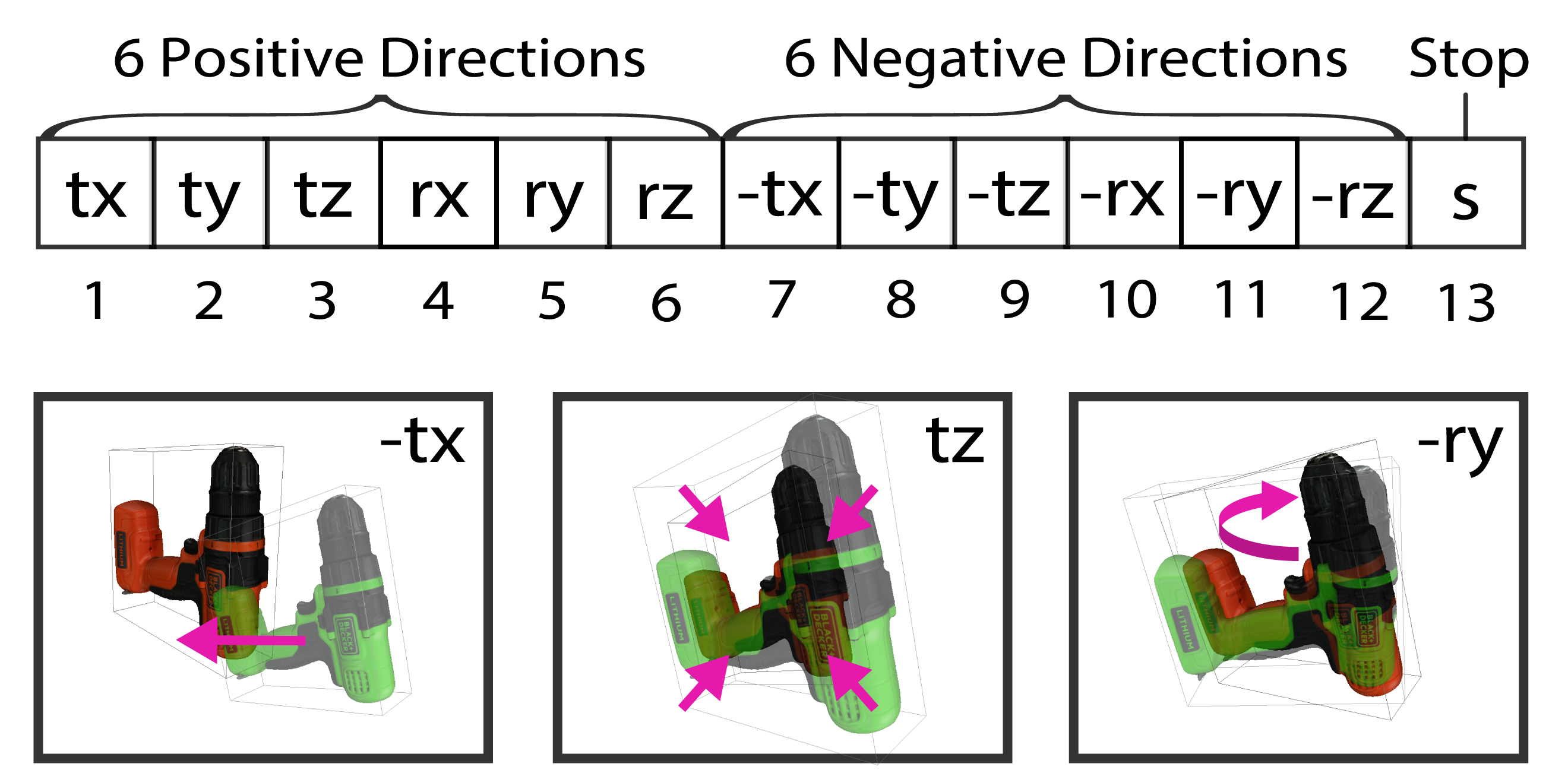}
    \caption{\textbf{Pose Update Actions.} One of 13 possible actions (top) determines the incremental rendering update (bottom).}
    \label{fig:actions}
\end{figure}
\begin{figure*}[t]
    \centering
    \includegraphics[width=.7\textwidth,clip]{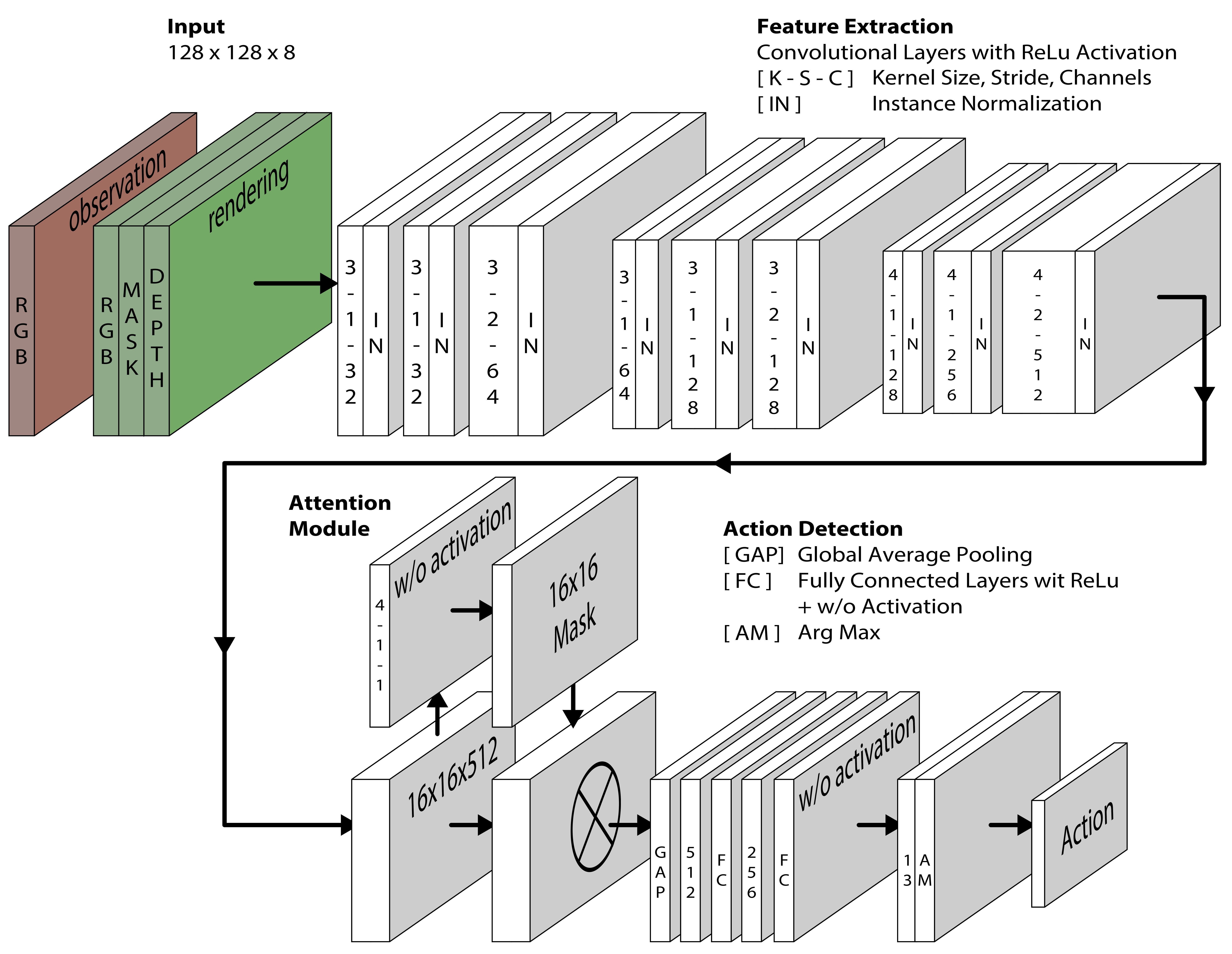}
    \caption{\textbf{Architecture Overview.} The input is the RGB video frame cropped and concatenated with the rendered RGB, rendered depth and rendered segmentation mask. A series of convolutional layers with activations are used to extract an embedding. An unsupervised attention mask is concatenated with it before an global average pooling layer. Two fully connected layers extract the set of action logits from which the most probable is selected with argmax.}
    \label{fig:architecture}
\end{figure*}
\section{Method}
\label{sec:method}\noindent
Our target is to optimize an action decision CNN to decide for iterative discrete actions to move a rendered 3D model to the observed position of the according object in an image sequence as shown in Fig.~\ref{fig:teaser}.
An initial pose is used to crop the image with the projected bounding box of the object.
We discretize the set of possible actions to move or not to move a 3D object depending on the six degrees of freedom for rigid displacement in space.
The 13 possible actions are divided into six pose actions for positive parameter adjustment, six for negative changes and an action to stop the process (i.e. not to move the object).
An action vector reads as
\begin{align}
    \text{a} = \left[ t_x, t_y, t_z, r_x, r_y, r_z \right] \times s \in \{ -1, 0, 1\}^6 \times \{ 0, 1\},
\end{align}
where each action is associated with a positive or negative direction.
The stop action is $s = 1$ while $s = 0$ indicates process continuation.
Associating all possible discrete actions with an element, we can describe the action vector also by $\text{v} \in \{ 0, 1 \}^{13}$ as shown in Fig.~\ref{fig:actions}.
A sequence $\text{v}_n, n \in \{1, \ldots N\}$ determines a result
\begin{align}
    \text{V} = \sum_{i=1}^{N} \text{v}_i = \sum_{i=1}^{N}\sum_{j=1}^{13} \text{v}_i \odot \text{e}_j = \sum_{i,j} \text{v}_{ij},
\end{align}
where one can split the action vector at step $i$ into its decision components $\text{v}_{ij}$ by element-wise multiplication with the discrete basis vector $\text{e}_j$ whose entries are $\delta_{jk}$ with $k \in \{1, \ldots 13\}$.
We aim to learn the next best decision $\text{v}_{ij}$ and iteratively sum these decisions where we can leverage an intermediate rendering of $\sum_{i,j} \text{v}_{ij}$ to compare with the currently observed image.\\
A synthetic dataset provides training images with the best decision defined as the one that minimizes the pose error
\begin{align}
     \text{v}_{\text{gt}} = \underset{\text{e}_j}{\argmin} \left\lVert \text{P}_{\text{gt}} - \text{P} \left( \sum_{i=1}^{m} \text{v}_i + \text{v}_{m+1} \odot \text{e}_j\right) \right\rVert
\end{align}
for a pose $\text{P}$.
This serves as supervision to train with the loss
\begin{align}
     \mathcal{L} = \| \text{v} - \text{v}_{\text{gt}} \|
\end{align}
minimizing the Hamming distance between the predicted action vector and the best decision.\\
For each of these actions, we set units depending on an image and a current crop: While movements for $t_x$, $t_y$ are measured in pixels and determine movements of the bounding box as such, $r_x$, $r_y$, $r_z$ are measured in degrees and $t_z$ is determined as the diameter in pixels of the current bounding box.
While an action can change the position and size of the crop, the image crop is always rescaled to a quadratic $n \times n$ patch of the same size as the rendering.

\noindent
We decide to implement the action decision CNN with a lightweight architecture that allows for training on a consumer laptop.
The architecture details are shown in Fig.~\ref{fig:architecture}.
An attention mechanism is implemented as guidance for the network to focus on relevant image regions and ignore occlusions.
This attention map is learnt in an unsupervised way during training to mask the embedded feature tensor and realize a weighted global average pooling.

\noindent
\textbf{Stopping Criteria \& Tracking Mode. }\noindent
Usually the iteration process is stopped with the stop action in frame $t$ and the last pose is used to initialize the process in frame $t+1$.
As we discretize the pose steps, the stop criterion, however, may not always be hit perfectly.
Moreover, the decision boundary between the stop criterion and some close action may lead to oscillations between two or multiple predictions close to the correct result.
To cope with this in practice, we can also stop the process early if we encounter oscillations and if an intermediate pose has been predicted before in the same loop or if a maximum number of iterations is reached.

\noindent
\textbf{Initialization \& Detection. }\noindent
We observe that the model tends to first align the rendering for the translation and performs rotation actions afterwards.
We make use of this observation and run the network with multiple seeds as a pose detection pipeline omitting the use of another model as necessary for~\cite{manhardt2018deep, li2018deepim, garon2018framework}.
For this, we randomly chose an object pose and seed the image at different locations by changing $t_x$ and $t_y$ for the pose.
We then run one iteration of the network in every location and record the values for $t_x$ and $t_y$.
We normalize the 2-vector given these inputs and generate a sparse vector field $\textbf{V}$ on top of the image as shown in Fig.~\ref{fig:detect} where we place these vectors at the seed centres.
This vector field is rather random for non-overlapping regions while its flux points toward the projection centre of the object if visible.
Applying a divergence operation $\text{W} = \nabla \cdot \textbf{V}$ on the smoothed vectors allows to find the object centre as the maximum of $\text{W}$ while analyzing $\text{W}$ helps also to determine a valuable bounding box size for a first crop.
Running the method on a coarsely discretized rotation space in this crop allows to find an initial rotation where the minimum number of iteration positively correlates with a possible starting rotation.
As the initial seeds can be calculated independent from each other, this process can be parallelized.


\section{Experiments}
\label{sec:experiments}\noindent
We implemented the model using the 3D renderer from unity~\cite{unitygameengine} with a customized version of the ML-agent toolkit~\cite{juliani2018unity}.
We combined it with TensorFlow
and train with a batch size of 32 using the ADAM~\cite{kingma:adam} optimizer with a learning rate of $10^{-4}$ and exponential decay of 5\% every 1k iterations.
We trained all our models until convergence (i.e. 25k iterations for object-specific training and 50k for multi-object training).
For all our experiments as well as training and dataset creation, we used a consumer laptop with an Intel Xeon E3-1505Mv6 CPU and an Nvdia Quadro P5000 mobile GPU.

\subsection{Training on Synthetic Data}
\label{subsec:dataset}
\begin{figure}[t]
    \centering
    \includegraphics[width=\linewidth,clip]{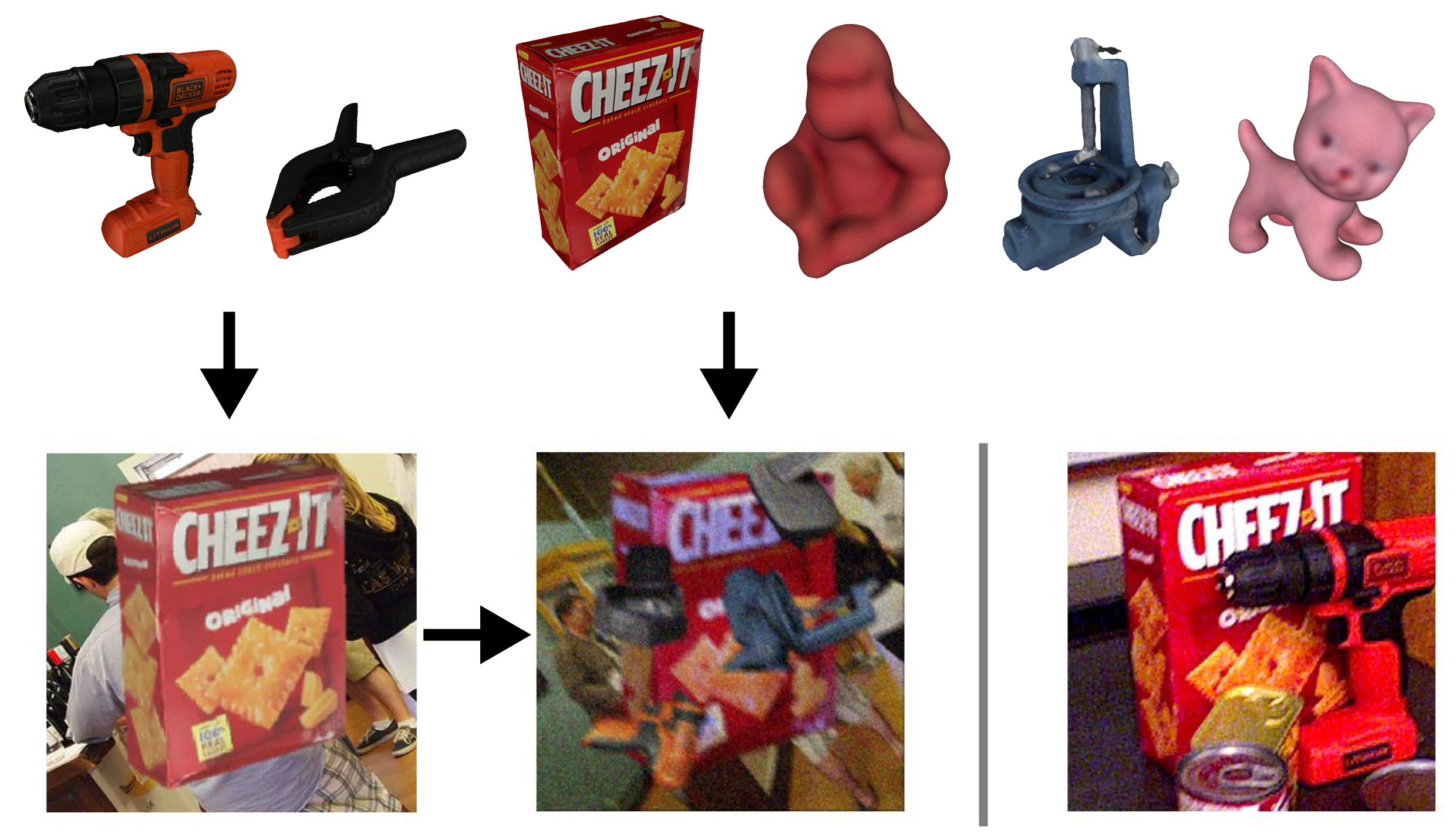}
    \caption{\textbf{Dataset Creation.} Synthetic 3D models are rendered in various poses on top of real 2D images. Augmentation in form of blur, light changes and occlusions is added. A comparison image from the real dataset is shown on the right.}
    \label{fig:dataset}
\end{figure}

\noindent
We propose a synthetic dataset generation pipeline where we render the 3D models with changing backgrounds and varying poses in clutter and occlusion on top of real images.
Following~\cite{kehl2017ssd} we use images from MS COCO~\cite{lin2014microsoft} as background.
We randomly pick 40k images from~\cite{lin2014microsoft} and use the high quality 3D models from YCB~\cite{xiang2018posecnn} and the models from Linemod~\cite{hinterstoisser2012} to render the objects during training in various poses as shown in Fig.~\ref{fig:dataset}.\\

\noindent
\textbf{Data Augmentation. }\noindent
We augment the renderings in different ways with occluders, crops, image blur as well as material and light changes before placing it on top of the real images.
As our network operates on cropped image patches of size $128 \times 128$ pixels, we perform the augmentation on these patches, too.
We synthetically generate 50k images for each YCB~\cite{xiang2018posecnn} object and 50k images for each Linemod~\cite{hinterstoisser2012} model.
The augmentation pipeline is described in detail in the supplementary material.
We consider these images as our synthetic ground truth.\\
\noindent
To simulate also the initial pose seeds, we produce a variety of 3D renderings without any augmentation a set of actions away from the related synthetic ground truth patch.
We want our method to work particularly well close to the correct result where it is crucial to take the right decisions in order to converge.
For this reason instead of rendering random seeds evenly distributed in pose space, we pay close attention near the ground truth by providing more training data in this region.
We group the pose seeds in five clusters: 10k each for YCB and Linemod.
The first cluster contains \textit{small} misalignment in only one action direction, where each action has an equal chance of $1/13$ to be picked, also the stop-action.
For the step size it holds $t_x, t_y \in \left[ 1, 5  \right]$, $t_z, r_i \in \left[ 1, 4  \right] \forall i$.
The second group consists of \textit{larger} misalignment in only one direction with equal chance.
For this we chose $t_x, t_y \in \left[ 5, 30  \right]$, $t_z \in \left[ 1, 15  \right]$, $r_i \in \left[ 4, 20  \right] \forall i$.
The third group is \textit{mixed} where we have one larger misalignment in one direction and the remaining actions are random small misalignment (e.g. $t_x = 10$ and all other directions are randomly chosen as in group one).
The fourth and fifth groups are a \textit{random small} and a \textit{random large} mix of misalignments from groups one and two.\\

\begin{figure}[t]
    \centering
    \includegraphics[width=\linewidth,clip]{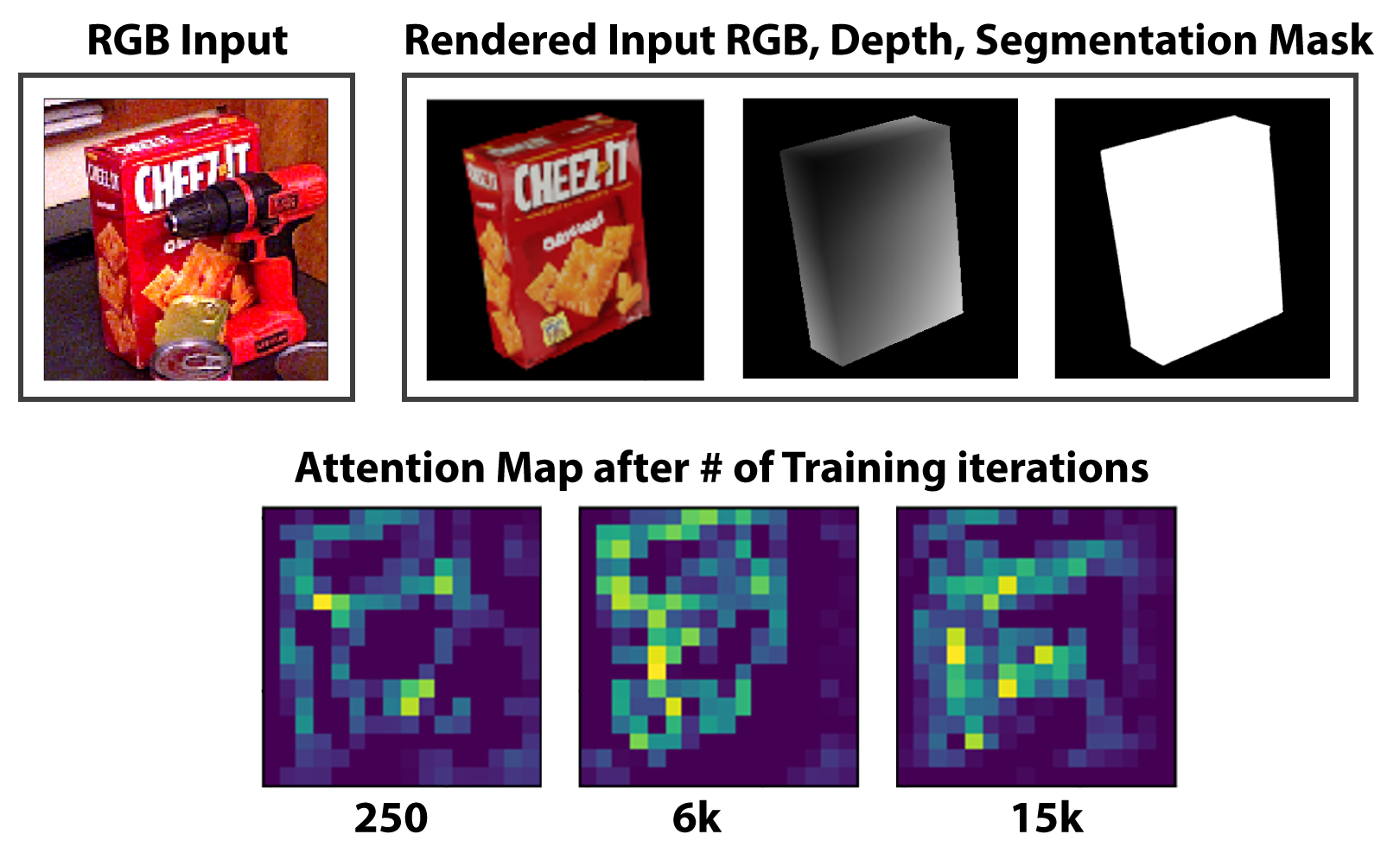}
    \caption{\textbf{Unsupervised Training of Attention Map.} The input RGB and input renderings are shown together with the results of the unsupervised training of the attention map after different numbers of training steps.}
    \label{fig:mask}
\end{figure}

\noindent\textbf{Training. }
We train networks for each YCB~\cite{xiang2018posecnn} model (object-specific training) and one network with mixed training including all YCB and Linemod models (multi-object training).
Fig.~\ref{fig:mask} shows the unsupervised training of our attention map on the same image after different number of iterations for training with cracker box.
It can be seen, that after attention on high gradient object regions (250 iterations), the mask emphasizes on the overall object geometry excluding big occlusion patches (6k iterations) before it learns to exclude the finer occluder details such as the front part of the drill (15k iterations).

\subsection{Pose Estimation \& Dataset Quality}
\label{subsec:quality}\noindent
\textbf{Datasets. }
High quality pose annotations are usually acquired with fiducial markers, manual annotation or a combination of both~\cite{hinterstoisser2012,brachmann2014}.
This process is very time-consuming and video annotations for 6D pose estimation are difficult to retrieve.
In order to produce the marker-free video pose dataset YCB~\cite{xiang2018posecnn}, the authors manually annotated only the poses of all the objects in the first frame of a sequence.
The ground truth labels for the rest of the frames within the sequences are retrieved by camera trajectory estimation.
Larger frame sets are possible, however, the quality of the annotations can vary.
The Laval~\cite{garon2018framework} video dataset circumvents this issue through the use of a motion capture system and retroreflective markers attached to the objects.
We test our models on these two datasets and evaluate both quantitatively and qualitatively.
The models in the YCB dataset are part of our training, while the objects from Laval are entirely \textit{unseen}.\\

\begin{figure}[t]
    \centering
    \includegraphics[width=\linewidth,clip]{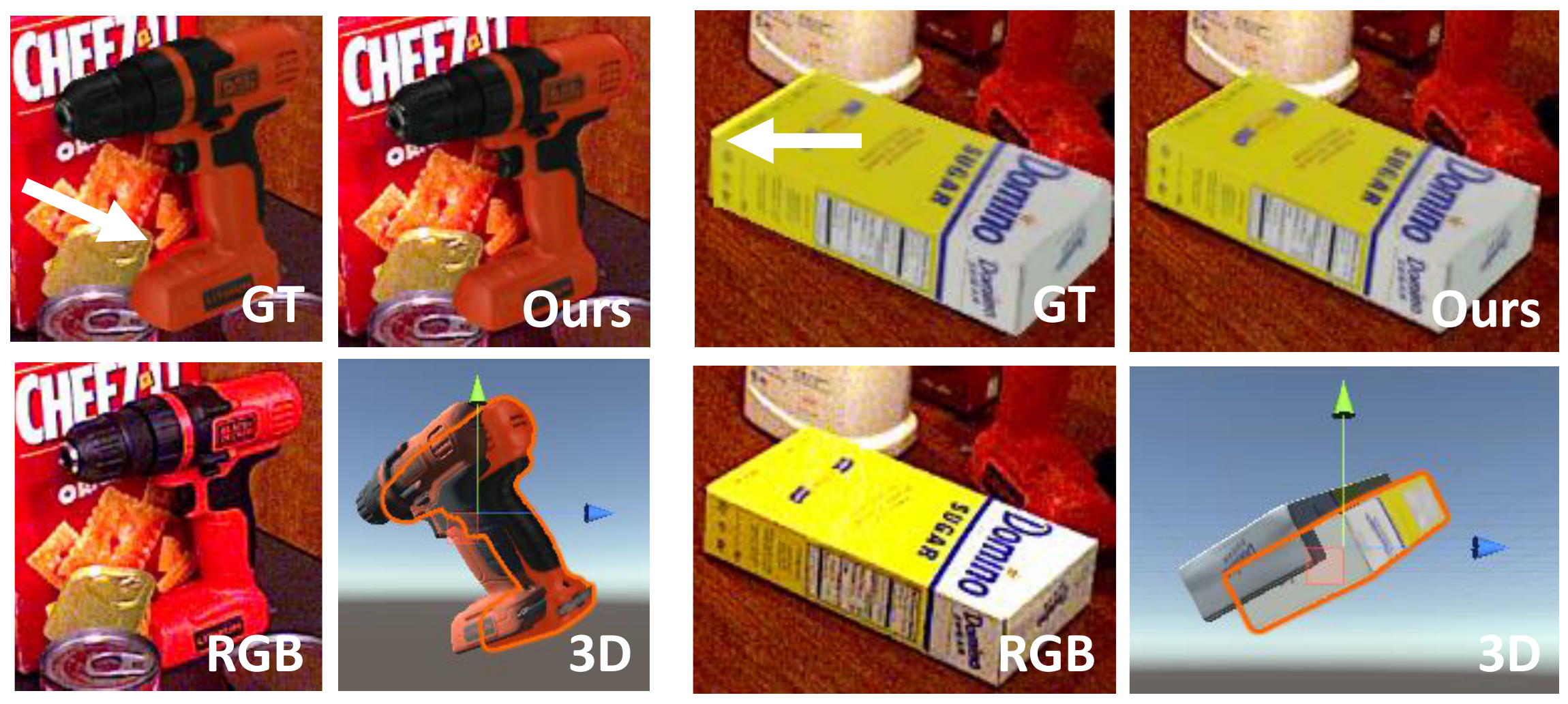}
    \caption{\textbf{Annotation Quality for YCB.} The input image is shown together with our prediction and the ground truth annotations. Arrows and 3D visualization are added to detail the difference in these cases where our estimation is considered incorrect.}
    \label{fig:ycb_error}
\end{figure}

\noindent
\textbf{Quantitative \& Qualitative Evaluation. }
For all quantitative experiments, we follow the protocol of~\cite{garon2017deep,garon2018framework} and reset the pose estimation with the annotated pose every $15$ frames.
The maximum number of action steps per frame is set to $30$.
At first, we test our networks trained on individual YCB models and compare with their ground truth poses~\cite{xiang2018posecnn}.
The result is reported in comparison with the state-of-the-art~\cite{xiang2018posecnn,fu2019deephmap,hu2019segmentation,oberweger2018making} in Tab.~\ref{tab:ycb_full} column two to six.
We utilize the 3D metrics for ADD and ADI (for symmetric objects) relative to the object diameter as proposed in~\cite{hinterstoisser2012}.\\
We can note an average improvement of 9.94\% compared to \cite{oberweger2018making} for our method and investigated the failure cases.
While most of them seem visually plausible, we still observe a significant accuracy variance between the video sequences in YCB which we further analyzed.
It turns out that the annotations for some of the objects are slightly shifted as shown in Fig.~\ref{fig:ycb_error}.
Our method -- in contrast to others with which we compare in Tab.~\ref{tab:ycb_full} -- is fully trained on synthetic data.
Thus, we cannot learn an annotation offset during training time due to the fact that our training setup provides pixel-perfect ground truth.
Further investigations revealed that the ground truth annotation quality is a common issue amongst multiple videos sequences in this dataset.
We believe that the main source for this is a slightly incorrect annotation in the first frame that propagates through the whole sequence, as the manual label is only given in frame one~\cite{xiang2018posecnn}.
We correct this shift by a single, constant translation delta for each sequence and rerun the evaluation.
The results are shown in the last column of Tab.~\ref{tab:ycb_full}, where the accuracy of our method improves significantly to a margin of 28.64\% over the state-of-the-art.

\begin{table*}[htbp]
\centering
\footnotesize
\begin{tabular}{l|rrrrr|r} \toprule
Model & {PC~\cite{xiang2018posecnn}} & {HMP~\cite{fu2019deephmap}} & {SD~\cite{hu2019segmentation}} & {HM~\cite{oberweger2018making}} & {Ours OS} & {Ours + Shift} \\
    \midrule
		002\_master\_chef\_can	&3.60&40.10&33.00&\textbf{75.80}&7.70&91.88\\
		003\_cracker\_box		&25.10&69.50&46.60&86.20&\textbf{88.36}&97.76\\
		004\_sugar\_box			&40.30&49.70&\textbf{75.60}&67.70&58.35&91.95\\
		005\_tomato\_soup\_can	&25.50&36.10&\textbf{40.80}&38.10&38.23&57.99\\
		006\_mustard\_bottle	&61.90&57.90&70.60&\textbf{95.20}&87.74&98.49\\
		007\_tuna\_fish\_can	&11.40&9.80&18.10&5.83&\textbf{47.90}&52.89\\
		008\_pudding\_box		&14.50&67.20&12.20&\textbf{82.20}&58.68&76.00\\
		009\_gelatin\_box		&12.10&59.10&59.40&\textbf{87.80}&37.08&89.20\\
		010\_potted\_meat\_can	&18.90&42.00&33.30&\textbf{46.50}&45.99&60.61\\
		011\_banana				&30.30&19.30&16.60&30.80&\textbf{74.02}&90.43\\
		019\_pitcher\_base		&15.60&58.50&90.00&57.90&\textbf{99.40}&100.00\\
		021\_bleach\_cleanser	&21.20&69.40&70.90&73.30&\textbf{95.04}&95.30\\
		\textit{024\_bowl}				&12.10&27.70&30.50&36.90&\textbf{99.44}&99.44\\
		025\_mug				&5.20&12.90&40.70&17.50&\textbf{45.35}&76.59\\
		035\_power\_drill		&29.90&51.80&63.50&\textbf{78.80}&52.77&97.35\\
		\textit{036\_wood\_block}		&10.70&35.70&27.70&33.90&\textbf{52.28}&63.48\\
		037\_scissors			&2.20&2.10&17.10&43.10&\textbf{63.33}&81.11\\
		040\_large\_marker		&3.40&3.60&4.80&8.88&\textbf{39.53}&41.73\\
		\textit{051\_large\_clamp}		&28.50&11.20&25.60&50.10&\textbf{64.01}&82.83\\
		\textit{052\_extra\_large\_clamp}&19.60&30.90&8.80&32.50&\textbf{88.02}&91.37\\
		\textit{061\_foam\_brick}		&54.50&55.40&34.70&66.30&\textbf{80.83}&80.83\\
	\midrule
	    Average					&21.26&38.57&39.07&53.11&\textbf{63.05}&81.75\\
	\bottomrule
\end{tabular}
\caption{Evaluation on the YCB dataset with our object-specific models. We compare the percentage of frames for which the 3D AD\{D$\vert$I\} error is $< 10\%$ of the object diameter~\cite{hinterstoisser2012}. Symmetric objects are shown in italic letters.}
\label{tab:ycb_full}
\end{table*}

\noindent
\textbf{Generalization and Ablation Study. }
Given these problematic initial annotations, we refrain form further interpretation of the results and investigate another dataset~\cite{garon2018framework}.
To the best of our knowledge, we are the first RGB-only method to report object-specific results on the challenging sequences of Laval~\cite{garon2018framework} where we test the generalization capabilities of our multi-object model.
Please note that the objects of the dataset have not been seen during training.
The results are summarized in Tab.~\ref{tab:laval} and Fig.~\ref{fig:failure} shows an example scenario.
To ablate our small network with a single loss term, we also provide the corresponding result for a model trained without the synthetic depth input channel. 
We follow the evaluation protocol of~\cite{garon2018framework} and report separately the average error for translation and rotation.
Tab.~\ref{tab:laval} shows that our multi-object model generalizes well on this dataset where the ground truth is acquired with a professional tracking system.
Both models are able to track the unseen object in translation. While the full model provides close results both for translation and rotation, the ablated model focuses only on the translation component and predicts stop once the object centre is aligned with only weak corrections for the rotation.
Without the depth rendering, the rotational error is significantly larger.
Rendering the synthetic depth helps with respect to the rotational accuracy.
This can be explained by the fact that moving the object in a close proximity to the observation does not require detailed understanding of depth while rotating it correctly is more intricate.

\begin{table}[htbp]
	\centering
	\footnotesize
	\begin{tabular}{l>{\raggedleft\arraybackslash}p{0.5cm}>{\raggedleft\arraybackslash}p{0.5cm}>{\raggedleft\arraybackslash}p{0.5cm}|>{\raggedleft\arraybackslash}p{0.5cm}>{\raggedleft\arraybackslash}p{0.5cm}>{\raggedleft\arraybackslash}p{0.5cm}} \toprule
		&\multicolumn{3}{c|}{\textbf{Ours full}}
		&\multicolumn{3}{c}{\textbf{Ours w/o D}}\\ \midrule
		\textbf{Occlusion}&  0\%&  15\%&  30\%&  0\%&  15\%&  30\%\\ \midrule
		\multicolumn{4}{l}{\textbf{Turtle}} \\ \midrule
		T[mm]& 5.92&9.91&12.91& 5.53&6.37&16.14\\
		R[deg]& 7.09& 14.87&14.87&18.31&20.13&26.03\\ \midrule
		\multicolumn{4}{l}{\textbf{Walkman}} \\ \midrule
		T[mm] &8.74&18.93&31.98&11.63&15.63&20.12\\
		R[deg]&6.97&11.33&21.17&40.68&44.47&50.18\\ \midrule
	\end{tabular}
	\caption{Evaluation result on Laval dataset for different levels of noise. We compare the full model to a model without rendered depth input. More objects are investigated in the suppl. material.}
	\label{tab:laval}
\end{table}
\begin{figure}[t]
    \centering
    \includegraphics[width=.8\linewidth,clip]{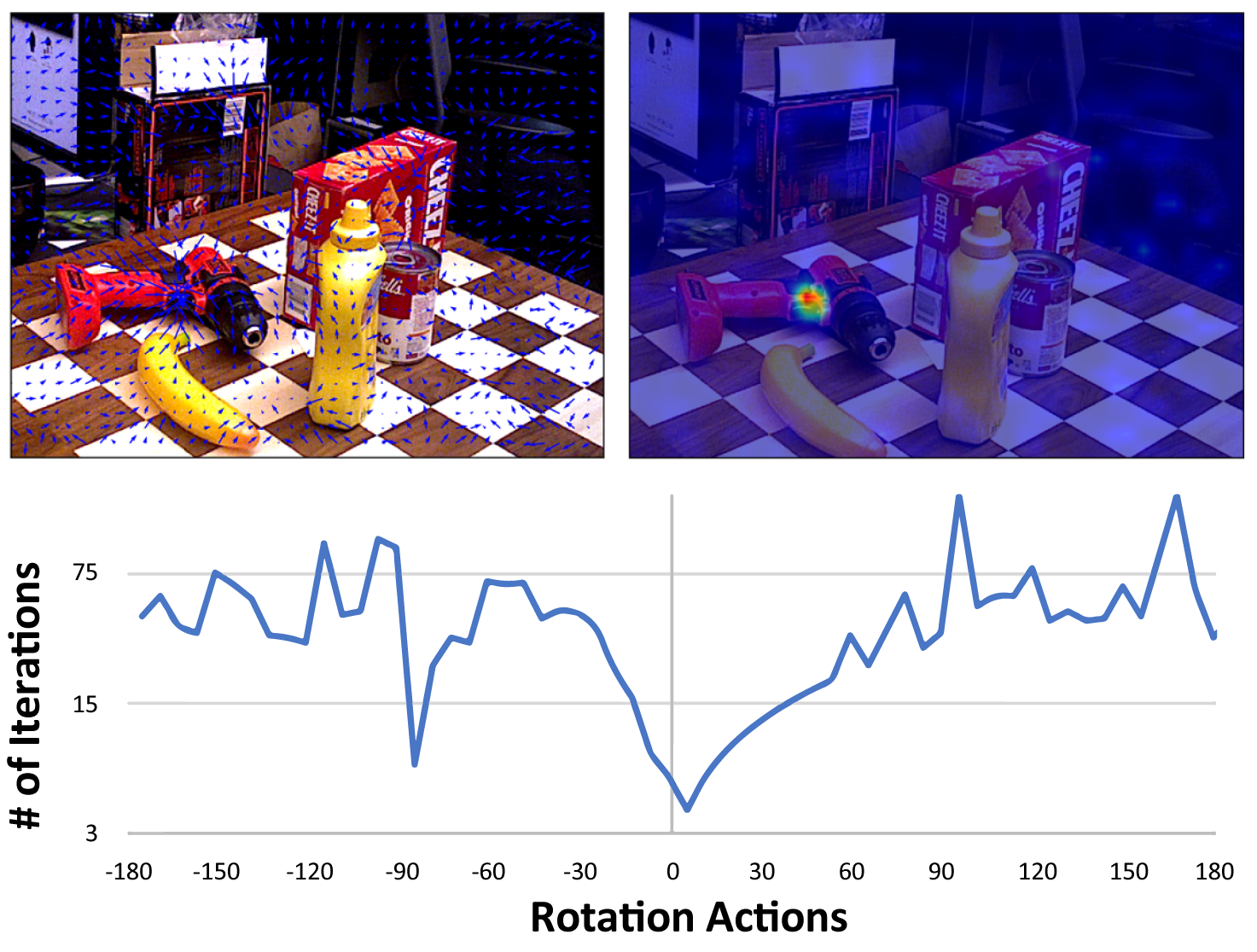}
    \caption{\textbf{Initial Point \& Rotation Seeding.} The predictions for $t_x$ and $t_y$ generate a vector field over the image (top left) whose divergence (top right) determines the initial point. Seeding a random rotation at this point allows to calculate the initial pose. The necessary number of iterations is plotted (bottom) against different seeds at a certain deviation from this rotation in just one action parameter (in this case $r_z$). A good initialization in the example is $+5$ actions away where the curve has its minimum.}
    \label{fig:detect}
\end{figure}

\subsection{Robustness \& Convergence}
\label{subsec:robustness}
\begin{figure}[t]
    \centering
    \includegraphics[width=\linewidth,clip]{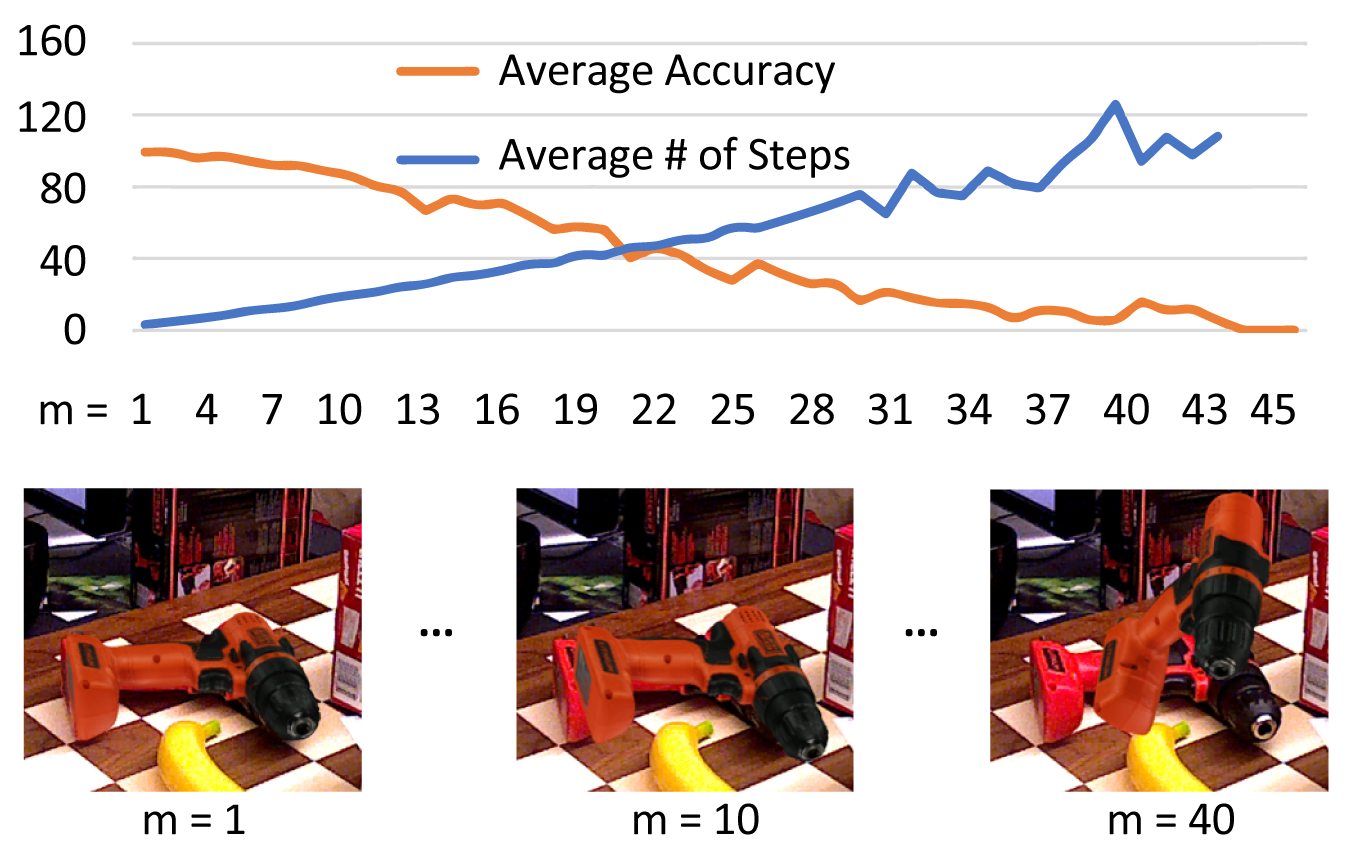}
    \caption{\textbf{Robustness Test.} The average ADD score is shown for increasing deviations (top) from the ground truth (orange) while the average number of steps the method needed for convergence is illustrated in blue. For deviations with $m \geq 43$ the method did not converge within 200 steps.}
    \label{fig:robustness}
\end{figure}

\noindent
The performance of conventional trackers largely depends on the difference between the correct pose and the initialization~\cite{akkaladevi2016tracking}.
As their paradigm is temporally consistent motion in videos, oftentimes close-to-correct poses are available from the result of the previous frame or from re-initialization with another algorithm~\cite{deng2019poserbpf} while we can use our framework for intrinsic initialization as shown in Fig.~\ref{fig:detect}.
Recent methods severely suffer if the initialization is too far off~\cite{garon2017deep,garon2018framework}.
Moreover, most conventional 3D trackers are not able to detect whether their estimation is correct or not.
In contrast to them, we propose a pipeline with a large convergence basin able to detect its own drift by analysing the number of steps and our stopping criterion.\\
\noindent
We test the convergence radius of our model by providing different initial poses with gradually increasing deviation from the correct result.
After manually checking the ground truth poses of the YCB dataset~\cite{xiang2018posecnn}, we decided to test with power drill on all keyframes from video sequence 50 which provides reliable annotations.
We prepare initial poses by deteriorating the ground truth annotations with increasing noise from the correct result to an initialization which is 270 actions apart.
This is done by adding actions to the GT pose with the state $\left[ t_x, t_y, t_z, r_x, r_y, r_z \right]$ in the form of:
\begin{equation}\label{eq:adding_noise}
     \Delta \cdot \left[ m(t_x), m(t_y), m(t_z), m(r_x), m(r_y), m(r_z) \right],\newline
\end{equation}
\begin{equation}
     \text{where} \quad m(q) = m \cdot \text{sgn} \left( X \right),
\end{equation}
for all state variables $q$.
We vary the value $m \in \{ 0, ..., 45 \}$ and $X$ is drawn from the uniform distribution $U(-1,1)$ and determines the direction of corruption.
The parameter $\Delta = 6$ sets the deviation for our test.\\
We use the individually trained model and set the stepsize for all actions to three.
Then we run the method and record the average ADD accuracy score as well as the average number of steps in case the model converges to the correct solution.
We randomly reduce the amount of keyframes for $m \in \{ 25, ..., 30 \}$ to 25\% and for $m \in \{ 31, ..., 45 \}$ to 10\% to avoid unreasonably long computations.
If convergence is not reached within 200 steps, we treat the run as a fail.
The results are summarized in Fig.~\ref{fig:robustness}.
Note that even for a large deviation of $m = 12$ which is significantly larger then the deviation found in the video sequence, our accuracy is $\text{ADD} = 73.8\%$.
Moreover, we can also see reasonable convergence in cases with 50\% or fewer bounding box overlap where other methods~\cite{garon2017deep} struggle and drift.\\
We use this wide convergence basin to show that our framework can be modified without retraining to also provide an initial pose close to the correct one in Fig.~\ref{fig:detect}.\\

\begin{figure}[t]
    \centering
    \includegraphics[width=\linewidth,clip]{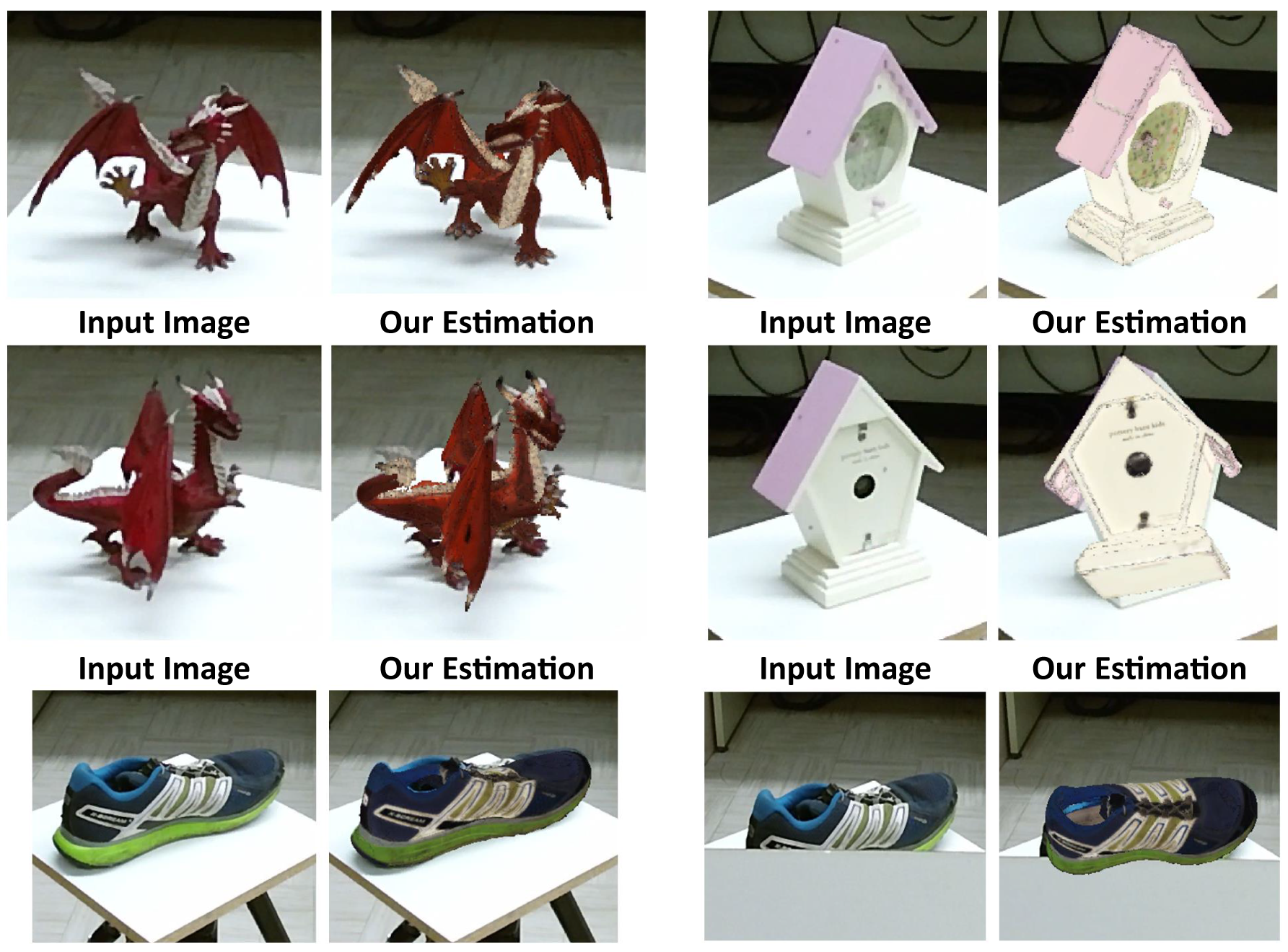}
    \caption{\textbf{Estimation Examples.} We show some prediction examples from the Laval~\cite{garon2018framework} dataset. Self-occluded fine details (Dragon), low texture (Clock) and occlusions (Shoe) can cause pose estimation failure for unseen objects.}
    \label{fig:failure}
\end{figure}

\noindent
\textbf{Failure Cases. }
Even though the convergence of our method is reliable, the network capacity is limited.
This results in pose estimation failures in case of heavy occlusions and fine detailed geometry.
Moreover, we share the issue with other RGB-only methods that low-textured objects are difficult to estimate reliably which results in drift in some cases as depicted in Fig.~\ref{fig:failure} together with further examples.\\

\noindent
\textbf{Runtime. } Due to the stop action, a dynamic runtime improvements can be reported depending on the motion present in the scene.
Since the number of iteration steps is non-static and the 3D rendering is negligible for this comparison, the overall runtime depends on two parameters: the action decision cycle and the number of actions.
In our implementation, the runtime for one loop in the cycle breaks down in the image preprocessing on CPU and the inference on the GPU.
We performe a runtime test averaging 512 iterations.
The results are shown in Tab.~\ref{tab:runtime}.

\begin{table}[htbp]
\centering
\footnotesize
\begin{tabular}{Sccccc} \toprule
        {Average Runtime on} & CPU & GPU & Total \\
    \midrule
        \text{Average Runtime in ms} & 14.6 & 5.2 & 19.8 \\
    \bottomrule
\end{tabular}
\caption{Average Runtime of Action Decision Process Cycle.}
\label{tab:runtime}
\end{table}

\noindent
Given the average of $4.2$~actions on our YCB tests, we report an overall average runtime of $83.16$~ms or $12$~FPS.
The runtime can be further increased if the image processing was also ported to the GPU.

\section{Conclusion}
\label{sec:conclusion}\noindent
We reformulated 6D pose estimation as an action decision process and presented a pipeline to solve it as a generic task without the need for object-specific training.
The method implements a dynamic runtime complexity depending on the inter-frame motion to increase performance and generalizes to unseen objects.
However, while improving the state-of-the-art for RGB-based video pose estimation, it still struggles in challenging cases for unseen objects.
Currently we search for the next best pose in every step.
An interesting direction for future research could be to integrate built up knowledge over time leveraging a more complex decision embedding space.

\clearpage
\section{Supplementary Material}
\label{sec:supplementary}\noindent
In this additional section, we detail the augmentation pipeline in Sec.~\ref{sec:augmentation} and report additional results on the YCB (Sec.~\ref{sec:ycb_auc}) and Laval (Sec.~\ref{sec:laval_additional}) datasets. Please also note the supplementary video that provides qualitative examples of our pipeline.\footnote{\href{http://campar.in.tum.de/Chair/PublicationDetail?pub=busam2020_moveIt}{Link to additional material.}}

\subsection{Data Augmentation Details}
\label{sec:augmentation}\noindent
We simulate two different blurs to augment the data with TensorFlow.
In 75\% of the cases, we randomly add motion blur and in 25\% of training scenarios a radial blur.
Both are generated with a mean of $\mu = 0$ and $\sigma = 0.05$ standard deviation for all three colour channels.
Variety in the exposures are augmented through changes of brightness, contrast and saturation values in the range of $\left[ 0.95, 1.25 \right]$.
For object material and light augmentation, we leverage the unity engine and simulate 20\% of unlit material and 80\% of standard material (i.e. metallic with $\left[ 0, 0.85  \right]$ and glossiness/smoothness with $\left[ 0, 0.8 \right]$).
Light is augmented with five point lights at random positions with an intensity drawn from $\left[ 0.5, 1.5  \right]$.
We change the light colour randomly by picking one colour from $C = \{ \text{blue, cyan, green, magenta, red, yellow, white} \}$ at every capture and set the same colour for all five lights.
The colour brightness for the light is randomly enhanced offering subtle additional variation in contrast to the intensity changes.
Then we randomly crop the rendering patch with $128 \times 128$ pixels to a height and width within $\left[ 96, 128  \right]$ and resize the patch to a value within $\left[ 32, 64  \right]$.
To simulate occlusion, we render 20k patches from YCB and Linemod models with random poses from which we pick four samples at each training step.
Firstly, they all are processed by the aforementioned blur and colour augmentation scheme.
In 50\% of the cases, we do not occlude the patch.
In the other cases we use these four samples for occlusion.
With a 12.5\% chance we respectively select either one, two or three occluders at random or use all four.
Finally, we crop the entire masked region of the augmentation pipeline in 25\% of the cases to simulate another occlusion scenario where we select the cropped region patch height and width randomly from $\left[ 72, 96  \right]$.
We apply this procedure to generate 50k images for each YCB~\cite{xiang2018posecnn} object and 50k images for each Linemod~\cite{hinterstoisser2012} model.

\subsection{Additional YCB Comparison}
\label{sec:ycb_auc}\noindent
The main paper shows a quantitative evaluation on the standard ADD metric~\cite{hinterstoisser2012} relative to the object diameter where a pose estimate is considered successful if its ADD value is below 10\% of the object diameter. The final ADD score is calculated by the percentage of frames with such a successful estimation.
Tables~\ref{tab:ycb_full_auc_1} and \ref{tab:ycb_full_auc_2} additionally compare the area under the ADD threshold curve (AUC) for varying absolute thresholds from zero to 0.1~m~\cite{xiang2018posecnn}.
The extensive study in comparison with the state-of-the-art shows that our method compares favourable on the standard benchmark (Ours OS) and significantly better with the shift-correction.

\begin{table*}[htbp]
\centering
\footnotesize
\begin{tabular}{l|rrrrrr|rr} \toprule
Model & {3DC~\cite{xiang2018posecnn}} & {PC~\cite{xiang2018posecnn}} & {CPC~\cite{capellen2019convposecnn}} & {PRBPF~\cite{deng2019poserbpf}} & {RKF~\cite{richter2019towards}} & {HM~\cite{oberweger2018making}}  & {Ours OS} & {+ Shift} \\
    \midrule
002\_master\_chef\_can		&	12.30	&	50.90	&	62.32	&	63.30	&	54.60	&	81.90	&	65.61	&	91.15	\\
003\_cracker\_box		&	16.80	&	51.70	&	66.69	&	77.80	&	57.60	&	83.60	&	84.34	&	90.74	\\
004\_sugar\_box		&	28.70	&	68.60	&	67.19	&	79.60	&	57.60	&	83.60	&	78.43	&	91.05	\\
005\_tomato\_soup\_can		&	27.30	&	66.00	&	75.52	&	73.00	&	68.30	&	79.80	&	66.83	&	76.06	\\
006\_mustard\_bottle		&	25.90	&	79.90	&	83.79	&	84.70	&	79.00	&	91.50	&	86.05	&	94.03	\\
007\_tuna\_fish\_can		&	5.40	&	70.40	&	60.98	&	64.20	&	43.50	&	48.70	&	65.90	&	69.12	\\
008\_pudding\_box		&	14.90	&	62.90	&	62.17	&	64.50	&	65.90	&	69.12	&	79.00	&	83.01	\\
009\_gelatin\_box		&	25.40	&	75.20	&	83.84	&	83.00	&	74.80	&	93.70	&	82.92	&	92.78	\\
010\_potted\_meat\_can		&	18.70	&	59.60	&	65.86	&	51.80	&	50.30	&	79.10	&	75.21	&	79.44	\\
011\_banana		&	3.20	&	72.30	&	37.74	&	18.40	&	8.20	&	51.70	&	84.99	&	90.19	\\
019\_pitcher\_base		&	27.30	&	52.50	&	62.19	&	63.70	&	77.80	&	69.40	&	85.14	&	94.22	\\
021\_bleach\_cleanser		&	25.20	&	50.50	&	55.14	&	60.50	&	59.30	&	76.20	&	89.27	&	90.68	\\
\textit{024\_bowl}		&	2.70	&	6.50	&	3.55	&	28.40	&	-	&	3.60	&	85.89	&	87.03	\\
025\_mug		&	9.00	&	57.70	&	45.83	&	77.90	&	69.10	&	53.90	&	78.95	&	87.83	\\
035\_power\_drill		&	18.00	&	55.10	&	76.47	&	71.80	&	71.40	&	82.90	&	76.56	&	91.95	\\
\textit{036\_wood\_block}		&	1.20	&	31.80	&	0.12	&	2.30	&	-	&	0.00	&	48.62	&	53.52	\\
037\_scissors		&	1.00	&	35.80	&	56.42	&	38.70	&	-	&	65.30	&	79.78	&	83.99	\\
040\_large\_marker		&	0.20	&	58.00	&	55.26	&	67.10	&	-	&	56.50	&	73.27	&	75.31	\\
\textit{051\_large\_clamp}		&	6.90	&	25.00	&	29.73	&	38.30	&	-	&	57.20	&	56.09	&	65.97	\\
\textit{052\_extra\_large\_clamp}		&	2.70	&	15.80	&	21.99	&	32.30	&	-	&	23.60	&	67.31	&	78.06	\\
\textit{061\_foam\_brick}		&	0.60	&	40.40	&	51.80	&	84.10	&	-	&	32.10	&	86.52	&	86.70	\\
\midrule			
Average		&	13.02	&	51.74	&	53.55	&	58.35	&	60.59	&	62.05	&	76.03	&	83.47	\\
    \bottomrule
\end{tabular}
\caption{Evaluation on the YCB dataset with our object-specific models. We compare the area under the ADD threshold curve (AUC) for varying thresholds from zero to 0.1~m. Symmetric objects are shown in italic letters.}
\label{tab:ycb_full_auc_1}
\end{table*}

\begin{table*}[htbp]
\centering
\footnotesize
\begin{tabular}{l|rrrrrr|rr} \toprule
Model & {R\&C~\cite{periyasamy2019refining}} & {Dope~\cite{tremblay2018deep}} &{HMP~\cite{fu2019deephmap}} & {MT~\cite{wang2019multi}} & {D-IM~\cite{li2018deepim}} & {PV-N~\cite{peng2019pvnet}} & {Ours OS} & {+ Shift} \\
    \midrule
002\_master\_chef\_can		&	76.70	&	-	&	75.80	&	62.70	&	71.20	&	81.60	&	65.61	&	91.15	\\
003\_cracker\_box		&	82.90	&	55.90	&	78.00	&	80.90	&	83.60	&	80.50	&	84.34	&	90.74	\\
004\_sugar\_box		&	86.40	&	75.70	&	76.50	&	83.80	&	94.10	&	84.90	&	78.43	&	91.05	\\
005\_tomato\_soup\_can		&	57.40	&	76.10	&	72.10	&	60.40	&	86.10	&	78.20	&	66.83	&	76.06	\\
006\_mustard\_bottle		&	86.70	&	81.90	&	78.90	&	85.10	&	91.50	&	88.30	&	86.05	&	94.03	\\
007\_tuna\_fish\_can		&	69.70	&	-	&	51.60	&	75.40	&	87.70	&	62.20	&	50.30	&	90.20   \\
008\_pudding\_box		&	68.80	&	-	&	85.60	&	17.70	&	82.70	&	85.20	&	79.00	&	83.01	\\
009\_gelatin\_box		&	73.00	&	-	&	86.70	&	79.90	&	91.90	&	88.70	&	82.92	&	92.78	\\
010\_potted\_meat\_can		&	74.60	&	39.40	&	70.10	&	55.00	&	76.20	&	65.10	&	75.21	&	79.44	\\
011\_banana		&	68.80	&	-	&	47.90	&	59.60	&	81.20	&	51.80	&	84.99	&	90.19	\\
019\_pitcher\_base		&	83.80	&	-	&	71.80	&	96.10	&	90.10	&	91.20	&	85.14	&	94.22	\\
021\_bleach\_cleanser		&	78.30	&	-	&	69.10	&	89.40	&	81.20	&	74.80	&	89.27	&	90.68	\\
\textit{024\_bowl}		&	1.50	&	-	&	-	&	49.50	&	8.60	&	-	&	85.89	&	87.03	\\
025\_mug		&	57.90	&	-	&	43.40	&	87.70	&	81.40	&	81.50	&	78.95	&	87.83	\\
035\_power\_drill		&	81.50	&	-	&	76.80	&	96.40	&	85.50	&	83.40	&	76.56	&	91.95	\\
\textit{036\_wood\_block}		&	0.00	&	-	&	-	&	43.80	&	60.00	&	-	&	48.62	&	53.52	\\
037\_scissors		&	75.40	&	-	&	42.90	&	60.20	&	60.90	&	54.80	&	79.78	&	83.99	\\
040\_large\_marker		&	59.80	&	-	&	47.60	&	87.50	&	75.60	&	35.80	&	73.27	&	75.31	\\
\textit{051\_large\_clamp}		&	75.30	&	-	&	-	&	90.70	&	48.40	&	-	&	56.09	&	65.97	\\
\textit{052\_extra\_large\_clamp}		&	20.40	&	-	&	-	&	88.10	&	31.00	&	-	&	67.31	&	78.06	\\
\textit{061\_foam\_brick}		&	37.00	&	-	&	-	&	26.30	&	35.90	&	-	&	86.52	&	86.70	\\
\midrule		
Average		&	62.66	&	65.80	&	67.18	&	70.30	&	71.66	&	74.25	&	76.03	&	83.47	\\
    \bottomrule
\end{tabular}
\caption{Evaluation on the YCB dataset with our object-specific models. We compare the area under the ADD threshold curve (AUC) for varying thresholds from zero to 0.1~m. Symmetric objects are shown in italic letters.}
\label{tab:ycb_full_auc_2}
\end{table*}

\subsection{Additional Laval Results}
\label{sec:laval_additional}\noindent
The main paper shows the results for the error on the Laval dataset~\cite{garon2018framework} for two objects. Table~\ref{tab:laval_additional} shows additional occlusion levels and the results for the remaining objects of the dataset which further support the claims from the main paper.

\begin{table*}[htbp]
	\centering
	\footnotesize
    \begin{tabular}{l>{\raggedleft\arraybackslash}p{1cm}>{\raggedleft\arraybackslash}p{1cm}>{\raggedleft\arraybackslash}p{1cm}>{\raggedleft\arraybackslash}p{1cm}|>{\raggedleft\arraybackslash}p{1cm}>{\raggedleft\arraybackslash}p{1cm}>{\raggedleft\arraybackslash}p{1cm}>{\raggedleft\arraybackslash}p{1cm}} \toprule
		&\multicolumn{4}{c|}{\textbf{Ours full}}
		&\multicolumn{4}{c}{\textbf{Ours w/o D}}\\ \midrule
		\textbf{Occlusion}&  0\%&  15\%&  30\%&  45\%&   0\%&  15\%&  30\%&  45\%\\ \midrule
		\multicolumn{4}{l}{\textbf{Clock}} \\ \midrule
		T[mm] & 14.02&20.54&25.85&51.92& 9.39&9.96&32.58&15.91\\
		R[deg] & 9.40&10.84&12.74&17.05& 29.15&27.92&30.72&28.40\\ \midrule
		\multicolumn{4}{l}{\textbf{Cookie Jar}} \\ \midrule
		T[mm] & 3.82&5.99&9.52&15.18& 1.79&2.75&11.62&5.95\\
		R[deg]&6.48&17.82&18.22&15.89& 28.77&18.18&24.30&19.02\\ \midrule
		\multicolumn{4}{l}{\textbf{Dog}} \\ \midrule
		T[mm] &12.09&28.37&55.48&77.91& 6.10&10.76&33.89&15.62\\
		R[deg]&11.70&14.21&22.43&23.80& 20.75&26.81&24.22&22.53\\ \midrule
		\multicolumn{4}{l}{\textbf{Dragon}} \\ \midrule
		T[mm] &22.47&29.39&36.37&40.06& 25.69&25.13&27.71&30.65\\
		R[deg]& 3.34&4.89&11.65&13.39& 27.16&36.40&37.61&30.94\\ \midrule
		\multicolumn{4}{l}{\textbf{Shoe}}  \\ \midrule
		T[mm] &9.72&17.91&24.33&37.34& 44.61&19.90&38.04&41.90\\
		R[deg]& 5.84&9.26&17.89&16.91& 62.78&39.47&43.50&24.73\\ \midrule
		\multicolumn{4}{l}{\textbf{Turtle}} \\ \midrule
		T[mm]& 5.92&9.91&12.91&23.92& 5.53&6.37&16.14&12.63\\
		R[deg] & 7.09&14.87&14.87&14.11& 18.31&20.13&26.03&24.97\\ \midrule
		\multicolumn{4}{l}{\textbf{Walkman}} \\ \midrule
		T[mm] &8.74&18.93&31.98&45.13& 11.63&15.63&20.12&31.30\\
		R[deg] & 6.97&11.33&21.17&22.26 &40.68&44.47&50.18&45.14\\ \midrule
		\multicolumn{4}{l}{\textbf{Watering Can}} \\ \midrule
		T[mm] &14.67&21.66&18.68&33.26& 11.61&20.54&20.96&26.10\\
		R[deg]& 11.89&19.80&23.43&33.54& 38.89&40.85&36.30&35.23\\ \midrule
	\end{tabular}
	\caption{Evaluation error on Laval dataset for different levels of noise. We compare the full model to a model without rendered depth input.}
	\label{tab:laval_additional}
\end{table*}

{\small
\bibliographystyle{ieee_fullname}
\bibliography{literature}

\begin{thebibliography}{10}\itemsep=-1pt

\bibitem{akkaladevi2016tracking}
Sharath Akkaladevi, Martin Ankerl, Christoph Heindl, and Andreas Pichler.
\newblock Tracking multiple rigid symmetric and non-symmetric objects in
  real-time using depth data.
\newblock In {\em 2016 IEEE International Conference on Robotics and Automation
  (ICRA)}, pages 5644--5649. IEEE, 2016.

\bibitem{bay2006surf}
Herbert Bay, Tinne Tuytelaars, and Luc Van~Gool.
\newblock {Surf: Speeded up robust features}.
\newblock In {\em European conference on computer vision}, pages 404--417.
  Springer, 2006.

\bibitem{besl1992method}
Paul~J Besl and Neil~D McKay.
\newblock Method for registration of 3-d shapes.
\newblock In {\em Sensor fusion IV: control paradigms and data structures},
  volume 1611, pages 586--606. International Society for Optics and Photonics,
  1992.

\bibitem{birdal2016x}
Tolga Birdal, Ievgeniia Dobryden, and Slobodan Ilic.
\newblock {X-tag: A fiducial tag for flexible and accurate bundle adjustment}.
\newblock In {\em 3D Vision (3DV), 2016 Fourth International Conference on},
  pages 556--564. IEEE, 2016.

\bibitem{brachmann2014}
Eric Brachmann, Alexander Krull, Frank Michel, Stefan Gumhold, Jamie Shotton,
  and Carsten Rother.
\newblock {Learning 6d object pose estimation using 3d object coordinates}.
\newblock In {\em European conference on computer vision}, pages 536--551.
  Springer, 2014.

\bibitem{busam2015stereo}
Benjamin Busam, Marco Esposito, Simon Che'Rose, Nassir Navab, and Benjamin
  Frisch.
\newblock {A stereo vision approach for cooperative robotic movement therapy}.
\newblock In {\em Proceedings of the IEEE International Conference on Computer
  Vision Workshops}, pages 127--135, 2015.

\bibitem{busam2016}
Benjamin Busam, Marco Esposito, Benjamin Frisch, and Nassir Navab.
\newblock {Quaternionic Upsampling: Hyperspherical Techniques for 6 DoF Pose
  Tracking}.
\newblock In {\em 2016 Fourth International Conference on 3D Vision (3DV)},
  pages 629--638. IEEE, 10 2016.

\bibitem{busam2019sterefo}
Benjamin Busam, Matthieu Hog, Steven McDonagh, and Gregory Slabaugh.
\newblock Sterefo: Efficient image refocusing with stereo vision.
\newblock In {\em Proceedings of the IEEE International Conference on Computer
  Vision Workshops}, pages 0--0, 2019.

\bibitem{busam2018markerless}
Benjamin Busam, Patrick Ruhkamp, Salvatore Virga, Beatrice Lentes, Julia
  Rackerseder, Nassir Navab, and Christoph Hennersperger.
\newblock Markerless inside-out tracking for 3d ultrasound compounding.
\newblock In {\em Simulation, Image Processing, and Ultrasound Systems for
  Assisted Diagnosis and Navigation}, pages 56--64. Springer, 2018.

\bibitem{cai2013fast}
Hongping Cai, Tom{\'a}{\v{s}} Werner, and Ji{\v{r}}{\'\i} Matas.
\newblock Fast detection of multiple textureless 3-d objects.
\newblock In {\em International Conference on Computer Vision Systems}, pages
  103--112. Springer, 2013.

\bibitem{capellen2019convposecnn}
Catherine Capellen, Max Schwarz, and Sven Behnke.
\newblock Convposecnn: Dense convolutional 6d object pose estimation.
\newblock {\em arXiv preprint arXiv:1912.07333}, 2019.

\bibitem{crivellaro2015novel}
Alberto Crivellaro, Mahdi Rad, Yannick Verdie, Kwang Moo~Yi, Pascal Fua, and
  Vincent Lepetit.
\newblock A novel representation of parts for accurate 3d object detection and
  tracking in monocular images.
\newblock In {\em Proceedings of the IEEE international conference on computer
  vision}, pages 4391--4399, 2015.

\bibitem{deng2019poserbpf}
Xinke Deng, Arsalan Mousavian, Yu Xiang, Fei Xia, Timothy Bretl, and Dieter
  Fox.
\newblock Poserbpf: A rao-blackwellized particle filter for 6d object pose
  tracking.
\newblock {\em arXiv preprint arXiv:1905.09304}, 2019.

\bibitem{do2018deep}
Thanh-Toan Do, Ming Cai, Trung Pham, and Ian Reid.
\newblock {Deep-6DPose: Recovering 6D Object Pose from a Single RGB Image}.
\newblock {\em arXiv preprint arXiv:1802.10367}, 2018.

\bibitem{doumanoglou2016recovering}
Andreas Doumanoglou, Rigas Kouskouridas, Sotiris Malassiotis, and Tae-Kyun Kim.
\newblock Recovering 6d object pose and predicting next-best-view in the crowd.
\newblock In {\em Proceedings of the IEEE Conference on Computer Vision and
  Pattern Recognition}, pages 3583--3592, 2016.

\bibitem{drost2017introducing}
Bertram Drost, Markus Ulrich, Paul Bergmann, Philipp Hartinger, and Carsten
  Steger.
\newblock Introducing mvtec itodd-a dataset for 3d object recognition in
  industry.
\newblock In {\em Proceedings of the IEEE International Conference on Computer
  Vision}, pages 2200--2208, 2017.

\bibitem{esposito2015cooperative}
Marco Esposito, Benjamin Busam, Christoph Hennersperger, Julia Rackerseder, An
  Lu, Nassir Navab, and Benjamin Frisch.
\newblock {Cooperative robotic gamma imaging: Enhancing us-guided needle
  biopsy}.
\newblock In {\em International Conference on Medical Image Computing and
  Computer-Assisted Intervention}, pages 611--618. Springer, 2015.

\bibitem{esposito2016multimodal}
Marco Esposito, Benjamin Busam, Christoph Hennersperger, Julia Rackerseder,
  Nassir Navab, and Benjamin Frisch.
\newblock {Multimodal US--gamma imaging using collaborative robotics for cancer
  staging biopsies}.
\newblock {\em International journal of computer assisted radiology and
  surgery}, 11(9):1561--1571, 2016.

\bibitem{fiala2005artag}
Mark Fiala.
\newblock {ARTag, a fiducial marker system using digital techniques}.
\newblock In {\em Computer Vision and Pattern Recognition, 2005. CVPR 2005.
  IEEE Computer Society Conference on}, volume~2, pages 590--596. IEEE, 2005.

\bibitem{fu2019deephmap}
Mingliang Fu and Weijia Zhou.
\newblock Deephmap++: Combined projection grouping and correspondence learning
  for full dof pose estimation.
\newblock {\em Sensors}, 19(5):1032, 2019.

\bibitem{garon2017deep}
Mathieu Garon and Jean-Fran{\c{c}}ois Lalonde.
\newblock Deep 6-dof tracking.
\newblock {\em IEEE transactions on visualization and computer graphics},
  23(11):2410--2418, 2017.

\bibitem{garon2018framework}
Mathieu Garon, Denis Laurendeau, and Jean-Fran{\c{c}}ois Lalonde.
\newblock A framework for evaluating 6-dof object trackers.
\newblock In {\em Proceedings of the European Conference on Computer Vision
  (ECCV)}, pages 582--597, 2018.

\bibitem{garrido2014automatic}
Sergio Garrido-Jurado, Rafael Mu{\~{n}}oz-Salinas, Francisco~José
  Madrid-Cuevas, and Manuel~Jesús
  Mar{\textbackslash}'{\textbackslash}in-Jim{\'{e}}nez.
\newblock {Automatic generation and detection of highly reliable fiducial
  markers under occlusion}.
\newblock {\em Pattern Recognition}, 47(6):2280--2292, 2014.

\bibitem{held20123d}
Robert Held, Ankit Gupta, Brian Curless, and Maneesh Agrawala.
\newblock 3d puppetry: a kinect-based interface for 3d animation.
\newblock In {\em UIST}, pages 423--434. Citeseer, 2012.

\bibitem{hinterstoisser2012}
Stefan Hinterstoisser, Vincent Lepetit, Slobodan Ilic, Stefan Holzer, Gary
  Bradski, Kurt Konolige, and Nassir Navab.
\newblock {Model based training, detection and pose estimation of texture-less
  3d objects in heavily cluttered scenes}.
\newblock In {\em Asian conference on computer vision}, pages 548--562.
  Springer, 2012.

\bibitem{hodan2017t}
Tom{\'a}{\v{s}} Hodan, Pavel Haluza, {\v{S}}tep{\'a}n Obdr{\v{z}}{\'a}lek, Jiri
  Matas, Manolis Lourakis, and Xenophon Zabulis.
\newblock T-less: An rgb-d dataset for 6d pose estimation of texture-less
  objects.
\newblock In {\em 2017 IEEE Winter Conference on Applications of Computer
  Vision (WACV)}, pages 880--888. IEEE, 2017.

\bibitem{hodan2018bop}
Tomas Hodan, Frank Michel, Eric Brachmann, Wadim Kehl, Anders GlentBuch, Dirk
  Kraft, Bertram Drost, Joel Vidal, Stephan Ihrke, Xenophon Zabulis, et~al.
\newblock Bop: Benchmark for 6d object pose estimation.
\newblock In {\em Proceedings of the European Conference on Computer Vision
  (ECCV)}, pages 19--34, 2018.

\bibitem{hodavn2015detection}
Tom{\'a}{\v{s}} Hoda{\v{n}}, Xenophon Zabulis, Manolis Lourakis,
  {\v{S}}t{\v{e}}p{\'a}n Obdr{\v{z}}{\'a}lek, and Ji{\v{r}}{\'\i} Matas.
\newblock Detection and fine 3d pose estimation of texture-less objects in
  rgb-d images.
\newblock In {\em 2015 IEEE/RSJ International Conference on Intelligent Robots
  and Systems (IROS)}, pages 4421--4428. IEEE, 2015.

\bibitem{Occlusion2018}
Aleksander Holynski and Johannes Kopf.
\newblock Fast depth densification for occlusion-aware augmented reality.
\newblock In {\em SIGGRAPH Asia 2018 Technical Papers}, page 194. ACM, 2018.

\bibitem{hu2019segmentation}
Yinlin Hu, Joachim Hugonot, Pascal Fua, and Mathieu Salzmann.
\newblock Segmentation-driven 6d object pose estimation.
\newblock In {\em Proceedings of the IEEE Conference on Computer Vision and
  Pattern Recognition}, pages 3385--3394, 2019.

\bibitem{joseph2015versatile}
David Joseph~Tan, Federico Tombari, Slobodan Ilic, and Nassir Navab.
\newblock A versatile learning-based 3d temporal tracker: Scalable, robust,
  online.
\newblock In {\em Proceedings of the IEEE International Conference on Computer
  Vision}, pages 693--701, 2015.

\bibitem{juliani2018unity}
Arthur Juliani, Vincent-Pierre Berges, Esh Vckay, Yuan Gao, Hunter Henry,
  Marwan Mattar, and Danny Lange.
\newblock Unity: A general platform for intelligent agents.
\newblock {\em arXiv preprint arXiv:1809.02627}, 2018.

\bibitem{Kaskman_2019_ICCV_Workshops}
Roman Kaskman, Sergey Zakharov, Ivan Shugurov, and Slobodan Ilic.
\newblock Homebreweddb: Rgb-d dataset for 6d pose estimation of 3d objects.
\newblock In {\em The IEEE International Conference on Computer Vision (ICCV)
  Workshops}, Oct 2019.

\bibitem{kato1999marker}
Hirokazu Kato and Mark Billinghurst.
\newblock {Marker tracking and hmd calibration for a video-based augmented
  reality conferencing system}.
\newblock In {\em Augmented Reality, 1999.(IWAR'99) Proceedings. 2nd IEEE and
  ACM International Workshop on}, pages 85--94. IEEE, 1999.

\bibitem{kehl2017ssd}
Wadim Kehl, Fabian Manhardt, Federico Tombari, Slobodan Ilic, and Nassir Navab.
\newblock {SSD-6D: Making RGB-based 3D detection and 6D pose estimation great
  again}.
\newblock In {\em Proceedings of the International Conference on Computer
  Vision (ICCV 2017), Venice, Italy}, pages 22--29, 2017.

\bibitem{kehl2016deep}
Wadim Kehl, Fausto Milletari, Federico Tombari, Slobodan Ilic, and Nassir
  Navab.
\newblock {Deep learning of local RGB-D patches for 3D object detection and 6D
  pose estimation}.
\newblock In {\em European Conference on Computer Vision}, pages 205--220.
  Springer, 2016.

\bibitem{kehl2017real}
Wadim Kehl, Federico Tombari, Slobodan Ilic, and Nassir Navab.
\newblock Real-time 3d model tracking in color and depth on a single cpu core.
\newblock In {\em Proceedings of the IEEE Conference on Computer Vision and
  Pattern Recognition}, pages 745--753, 2017.

\bibitem{kehl2015hashmod}
Wadim Kehl, Federico Tombari, Nassir Navab, Slobodan Ilic, and Vincent Lepetit.
\newblock Hashmod: A hashing method for scalable 3d object detection.
\newblock In {\em BMVC}, volume~1, page~2, 2015.

\bibitem{kingma:adam}
Diederick~P Kingma and Jimmy Ba.
\newblock Adam: A method for stochastic optimization.
\newblock In {\em International Conference on Learning Representations (ICLR)},
  2015.

\bibitem{Knapitsch2017}
Arno Knapitsch, Jaesik Park, Qian-Yi Zhou, and Vladlen Koltun.
\newblock Tanks and temples: Benchmarking large-scale scene reconstruction.
\newblock {\em ACM Transactions on Graphics}, 36(4), 2017.

\bibitem{krull2017poseagent}
Alexander Krull, Eric Brachmann, Sebastian Nowozin, Frank Michel, Jamie
  Shotton, and Carsten Rother.
\newblock Poseagent: Budget-constrained 6d object pose estimation via
  reinforcement learning.
\newblock In {\em Proceedings of the IEEE Conference on Computer Vision and
  Pattern Recognition}, pages 6702--6710, 2017.

\bibitem{epnp_2009}
Vincent Lepetit, Francesc Moreno-Noguer, and Pascal Fua.
\newblock {EPnP: An Accurate O(n) Solution to the PnP Problem}.
\newblock {\em International Journal of Computer Vision}, 81(2):155--166, 2
  2009.

\bibitem{leutenegger2011brisk}
Stefan Leutenegger, Margarita Chli, and Roland~Y Siegwart.
\newblock {BRISK: Binary robust invariant scalable keypoints}.
\newblock In {\em Computer Vision (ICCV), 2011 IEEE International Conference
  on}, pages 2548--2555. IEEE, 2011.

\bibitem{li2010location}
Yunpeng Li, Noah Snavely, and Daniel~P Huttenlocher.
\newblock {Location recognition using prioritized feature matching}.
\newblock In {\em European conference on computer vision}, pages 791--804.
  Springer, 2010.

\bibitem{li2018deepim}
Yi Li, Gu Wang, Xiangyang Ji, Yu Xiang, and Dieter Fox.
\newblock Deepim: Deep iterative matching for 6d pose estimation.
\newblock In {\em Proceedings of the European Conference on Computer Vision
  (ECCV)}, pages 683--698, 2018.

\bibitem{Li_2019_ICCV}
Zhigang Li, Gu Wang, and Xiangyang Ji.
\newblock Cdpn: Coordinates-based disentangled pose network for real-time
  rgb-based 6-dof object pose estimation.
\newblock In {\em The IEEE International Conference on Computer Vision (ICCV)},
  October 2019.

\bibitem{lin2014microsoft}
Tsung-Yi Lin, Michael Maire, Serge Belongie, James Hays, Pietro Perona, Deva
  Ramanan, Piotr Doll{\'a}r, and C~Lawrence Zitnick.
\newblock Microsoft coco: Common objects in context.
\newblock In {\em European conference on computer vision}, pages 740--755.
  Springer, 2014.

\bibitem{lowe2004distinctive}
David~G Lowe.
\newblock {Distinctive image features from scale-invariant keypoints}.
\newblock {\em International journal of computer vision}, 60(2):91--110, 2004.

\bibitem{manhardt2019explaining}
Fabian Manhardt, Diego~Martin Arroyo, Christian Rupprecht, Benjamin Busam,
  Tolga Birdal, Nassir Navab, and Federico Tombari.
\newblock Explaining the ambiguity of object detection and 6d pose from visual
  data.
\newblock In {\em Proceedings of the IEEE International Conference on Computer
  Vision}, pages 6841--6850, 2019.

\bibitem{manhardt2018deep}
Fabian Manhardt, Wadim Kehl, Nassir Navab, and Federico Tombari.
\newblock Deep model-based 6d pose refinement in rgb.
\newblock In {\em Proceedings of the European Conference on Computer Vision
  (ECCV)}, pages 800--815, 2018.

\bibitem{manhardt2020cps}
Fabian Manhardt, Gu Wang, Benjamin Busam, Manuel Nickel, Sven Meier, Luca
  Minciullo, Xiangyang Ji, and Nassir Navab.
\newblock Cps++: Improving class-level 6d pose and shape estimation from
  monocular images with self-supervised learning, 2020.

\bibitem{mur2015orb}
Raul Mur-Artal, Jose Maria~Martinez Montiel, and Juan~D Tardos.
\newblock Orb-slam: a versatile and accurate monocular slam system.
\newblock {\em IEEE transactions on robotics}, 31(5):1147--1163, 2015.

\bibitem{mur2017orb}
Raul Mur-Artal and Juan~D Tard{\'o}s.
\newblock Orb-slam2: An open-source slam system for monocular, stereo, and
  rgb-d cameras.
\newblock {\em IEEE Transactions on Robotics}, 33(5):1255--1262, 2017.

\bibitem{oberweger2018making}
Markus Oberweger, Mahdi Rad, and Vincent Lepetit.
\newblock Making deep heatmaps robust to partial occlusions for 3d object pose
  estimation.
\newblock In {\em Proceedings of the European Conference on Computer Vision
  (ECCV)}, pages 119--134, 2018.

\bibitem{olson2011apriltag}
Edwin Olson.
\newblock {AprilTag: A robust and flexible visual fiducial system}.
\newblock In {\em Robotics and Automation (ICRA), 2011 IEEE International
  Conference on}, pages 3400--3407. IEEE, 2011.

\bibitem{pavlakos20176}
Georgios Pavlakos, Xiaowei Zhou, Aaron Chan, Konstantinos~G Derpanis, and
  Kostas Daniilidis.
\newblock {6-dof object pose from semantic keypoints}.
\newblock In {\em Robotics and Automation (ICRA), 2017 IEEE International
  Conference on}, pages 2011--2018. IEEE, 2017.

\bibitem{peng2019pvnet}
Sida Peng, Yuan Liu, Qixing Huang, Xiaowei Zhou, and Hujun Bao.
\newblock Pvnet: Pixel-wise voting network for 6dof pose estimation.
\newblock In {\em Proceedings of the IEEE Conference on Computer Vision and
  Pattern Recognition}, pages 4561--4570, 2019.

\bibitem{periyasamy2019refining}
Arul~Selvam Periyasamy, Max Schwarz, and Sven Behnke.
\newblock Refining 6d object pose predictions using abstract
  render-and-compare.
\newblock {\em arXiv preprint arXiv:1910.03412}, 2019.

\bibitem{pitteri2019cornet}
Giorgia Pitteri, Slobodan Ilic, and Vincent Lepetit.
\newblock Cornet: Generic 3d corners for 6d pose estimation of new objects
  without retraining.
\newblock In {\em Proceedings of the IEEE International Conference on Computer
  Vision Workshops}, pages 0--0, 2019.

\bibitem{rad2017bb8}
Mahdi Rad and Vincent Lepetit.
\newblock {BB8: A scalable, accurate, robust to partial occlusion method for
  predicting the 3D poses of challenging objects without using depth}.
\newblock In {\em ICCV}, 2017.

\bibitem{rad2018domain}
Mahdi Rad, Markus Oberweger, and Vincent Lepetit.
\newblock Domain transfer for 3d pose estimation from color images without
  manual annotations.
\newblock In {\em Asian Conference on Computer Vision}, pages 69--84. Springer,
  2018.

\bibitem{redmon2016you}
Joseph Redmon, Santosh Divvala, Ross Girshick, and Ali Farhadi.
\newblock {You only look once: Unified, real-time object detection}.
\newblock In {\em Proceedings of the IEEE conference on computer vision and
  pattern recognition}, pages 779--788, 2016.

\bibitem{rennie2016dataset}
Colin Rennie, Rahul Shome, Kostas~E Bekris, and Alberto~F De~Souza.
\newblock A dataset for improved rgbd-based object detection and pose
  estimation for warehouse pick-and-place.
\newblock {\em IEEE Robotics and Automation Letters}, 1(2):1179--1185, 2016.

\bibitem{richter2019towards}
Jesse Richter-Klug and Udo Frese.
\newblock Towards meaningful uncertainty information for cnn based 6d pose
  estimates.
\newblock In {\em International Conference on Computer Vision Systems}, pages
  408--422. Springer, 2019.

\bibitem{orb_2011}
Ethan Rublee, Vincent Rabaud, Kurt Konolige, and Gary Bradski.
\newblock {ORB: An efficient alternative to SIFT or SURF}.
\newblock In {\em 2011 International Conference on Computer Vision}, pages
  2564--2571. IEEE, 11 2011.

\bibitem{rusinkiewicz2001efficient}
Szymon Rusinkiewicz and Marc Levoy.
\newblock Efficient variants of the icp algorithm.
\newblock In {\em 3dim}, volume~1, pages 145--152, 2001.

\bibitem{segal2009generalized}
Aleksandr Segal, Dirk Haehnel, and Sebastian Thrun.
\newblock Generalized-icp.
\newblock In {\em Robotics: science and systems}, volume~2, page 435. Seattle,
  WA, 2009.

\bibitem{shao2020pfrl}
Jianzhun Shao, Yuhang Jiang, Gu Wang, Zhigang Li, and Xiangyang Ji.
\newblock Pfrl: Pose-free reinforcement learning for 6d pose estimation.
\newblock In {\em Proceedings of the IEEE/CVF Conference on Computer Vision and
  Pattern Recognition}, pages 11454--11463, 2020.

\bibitem{sundermeyer2018implicit}
Martin Sundermeyer, Zoltan-Csaba Marton, Maximilian Durner, Manuel Brucker, and
  Rudolph Triebel.
\newblock {Implicit 3D orientation learning for 6D object detection from RGB
  images}.
\newblock In {\em European Conference on Computer Vision}, pages 712--729.
  Springer, 2018.

\bibitem{Tan_2014_CVPR}
David~J. Tan and Slobodan Ilic.
\newblock Multi-forest tracker: A chameleon in tracking.
\newblock In {\em The IEEE Conference on Computer Vision and Pattern
  Recognition (CVPR)}, June 2014.

\bibitem{tejani2014latent}
Alykhan Tejani, Danhang Tang, Rigas Kouskouridas, and Tae-Kyun Kim.
\newblock {Latent-class hough forests for 3D object detection and pose
  estimation}.
\newblock In {\em European Conference on Computer Vision}, pages 462--477.
  Springer, 2014.

\bibitem{tekin2018real}
Bugra Tekin, Sudipta~N Sinha, and Pascal Fua.
\newblock {Real-time seamless single shot 6D object pose prediction}.
\newblock {\em CVPR}, 2018.

\bibitem{tremblay2018deep}
Jonathan Tremblay, Thang To, Balakumar Sundaralingam, Yu Xiang, Dieter Fox, and
  Stan Birchfield.
\newblock Deep object pose estimation for semantic robotic grasping of
  household objects.
\newblock {\em arXiv preprint arXiv:1809.10790}, 2018.

\bibitem{unitygameengine}
{Unity Technologies}.
\newblock Unity.
\newblock \url{https://unity3d.com/unity/whats-new/2018.3.6}.
\newblock Accessed: 2019-06-10.

\bibitem{wang2019densefusion}
Chen Wang, Danfei Xu, Yuke Zhu, Roberto Mart{\'\i}n-Mart{\'\i}n, Cewu Lu, Li
  Fei-Fei, and Silvio Savarese.
\newblock Densefusion: 6d object pose estimation by iterative dense fusion.
\newblock In {\em Proceedings of the IEEE Conference on Computer Vision and
  Pattern Recognition}, pages 3343--3352, 2019.

\bibitem{wang2019normalized}
He Wang, Srinath Sridhar, Jingwei Huang, Julien Valentin, Shuran Song, and
  Leonidas~J Guibas.
\newblock Normalized object coordinate space for category-level 6d object pose
  and size estimation.
\newblock In {\em Proceedings of the IEEE Conference on Computer Vision and
  Pattern Recognition}, pages 2642--2651, 2019.

\bibitem{wang2017gracker}
Tao Wang and Haibin Ling.
\newblock Gracker: A graph-based planar object tracker.
\newblock {\em IEEE transactions on pattern analysis and machine intelligence},
  40(6):1494--1501, 2017.

\bibitem{wang2019multi}
Yurui Wang, Shaokun Jin, and Yongsheng Ou.
\newblock A multi-task learning convolutional neural network for object pose
  estimation.
\newblock In {\em International Conference on Robotics and Biomimetics
  (ROBIO)}, pages 284--289. IEEE, 2019.

\bibitem{wohlhart2015learning}
Paul Wohlhart and Vincent Lepetit.
\newblock {Learning descriptors for object recognition and 3d pose estimation}.
\newblock In {\em Proceedings of the IEEE Conference on Computer Vision and
  Pattern Recognition}, pages 3109--3118, 2015.

\bibitem{wu20083d}
Changchang Wu, Friedrich Fraundorfer, Jan-Michael Frahm, and Marc Pollefeys.
\newblock {3D model search and pose estimation from single images using VIP
  features}.
\newblock In {\em Computer Vision and Pattern Recognition Workshops, 2008.
  CVPRW'08. IEEE Computer Society Conference on}, pages 1--8. IEEE, 2008.

\bibitem{xiang2018posecnn}
Yu Xiang, Tanner Schmidt, Venkatraman Narayanan, and Dieter Fox.
\newblock Posecnn: A convolutional neural network for 6d object pose estimation
  in cluttered scenes.
\newblock {\em Robotics: Science and Systems (RSS)}, 2018.

\bibitem{yuheng2013star3d}
Carl Yuheng~Ren, Victor Prisacariu, David Murray, and Ian Reid.
\newblock Star3d: Simultaneous tracking and reconstruction of 3d objects using
  rgb-d data.
\newblock In {\em Proceedings of the IEEE International Conference on Computer
  Vision}, pages 1561--1568, 2013.

\bibitem{Yun_2017_CVPR}
Sangdoo Yun, Jongwon Choi, Youngjoon Yoo, Kimin Yun, and Jin Young~Choi.
\newblock Action-decision networks for visual tracking with deep reinforcement
  learning.
\newblock In {\em The IEEE Conference on Computer Vision and Pattern
  Recognition (CVPR)}, July 2017.

\bibitem{Zakharov_2019_ICCV}
Sergey Zakharov, Ivan Shugurov, and Slobodan Ilic.
\newblock Dpod: 6d pose object detector and refiner.
\newblock In {\em The IEEE International Conference on Computer Vision (ICCV)},
  October 2019.

\bibitem{zhang1994iterative}
Zhengyou Zhang.
\newblock {Iterative point matching for registration of free-form curves and
  surfaces}.
\newblock {\em International journal of computer vision}, 13(2):119--152, 1994.

\bibitem{zhou2019continuity}
Yi Zhou, Connelly Barnes, Jingwan Lu, Jimei Yang, and Hao Li.
\newblock On the continuity of rotation representations in neural networks.
\newblock In {\em Proceedings of the IEEE Conference on Computer Vision and
  Pattern Recognition}, pages 5745--5753, 2019.

\end{thebibliography}
}


\end{document}